\documentclass{article} % For LaTeX2e
\usepackage{iclr2027_conference,times}

\usepackage{amsmath,amsfonts,bm}

\def\eqref#1{equation~\ref{#1}}
\def\1{\bm{1}}

\DeclareMathAlphabet{\mathsfit}{\encodingdefault}{\sfdefault}{m}{sl}
\SetMathAlphabet{\mathsfit}{bold}{\encodingdefault}{\sfdefault}{bx}{n}

\usepackage{hyperref}
\usepackage{url}
\usepackage{graphicx} % jhkim
\usepackage{wrapfig} % jhkim
\usepackage{booktabs}  % jhkim
\usepackage{graphicx}  % jhkim
\usepackage{makecell}  % jhkim
\usepackage{amssymb}  % jhkim
\usepackage{threeparttable}  % jhkim
\usepackage{multirow}
\usepackage{tabularray}
\UseTblrLibrary{booktabs}
\usepackage{capt-of}
\usepackage{titletoc}
\usepackage{subcaption}
\usepackage{xcolor}
\definecolor{darkbrown}{RGB}{110,55,30}
\definecolor{darkorange}{RGB}{173, 46, 35}
\definecolor{darkorange}{RGB}{191, 10, 15}

\makeatletter
\newcommand{\CenteredTwoRowHeader}[1]{%
  \begingroup
  \sbox0{\makecell[c]{\textbf{NAVSIM}\\\textbf{Open Loop}}}%
  \dimen0=\dp0
  \ifdim\dimen0<\dp\@arstrutbox
      \dimen0=\dp\@arstrutbox
  \fi
  \advance\dimen0 by
      \dimexpr(\aboverulesep+\cmidrulewidth+\belowrulesep)/2\relax
  \sbox2{\makecell[c]{#1}}%
  \dimen2=\dimexpr(\dp2-\ht2)/2-\dimen0\relax
  \raisebox{\dimen2}[0pt][0pt]{\usebox2}%
  \endgroup
}
\makeatother

\title{Odyssey: A Closed-Loop Benchmark for Long-Horizon Real-World Driving with Explicit Navigation Routes}

\author{%
  \makebox[\dimexpr\textwidth-2\tabcolsep\relax][c]{%
  \begin{tabular}[t]{@{}c@{}}
  Jungho Kim\thanks{Equal contribution.} \quad
  Hongjae Shin\footnotemark[1] \quad
  Seunghoon Yu\footnotemark[1] \quad
  Heecheol Yoo\footnotemark[1] \quad
  Myeongjun Kim\footnotemark[1] \quad
  Jiyong Oh\footnotemark[1] \\
  Donghyuk Kwak \quad
  Seunghyeop Nam \quad
  Haesung Oh \quad
  Hyunju Kim \quad
  Hyungchan Cho \\
  Jaehyun Park \quad
  Soo Won Seo \quad
  Jun Won Choi\thanks{Corresponding author.} \\[6pt]
  {\normalfont Seoul National University, Republic of Korea} \\[4pt]
{\normalfont\small
  \href{https://snu-adr.github.io/Odyssey-Page/}{%
  \textcolor{darkorange}{%
  \nolinkurl{https://snu-adr.github.io/Odyssey-Page/}}}}
  \end{tabular}%
  }%
  }

\iclrfinalcopy % Uncomment for camera-ready version, but NOT for submission.
\begin{document}

\maketitle
\fancyhead{}
\lhead{Preprint}

\begin{figure}[h]
\centering
\vspace{-20pt}
\includegraphics[width=0.95\columnwidth]{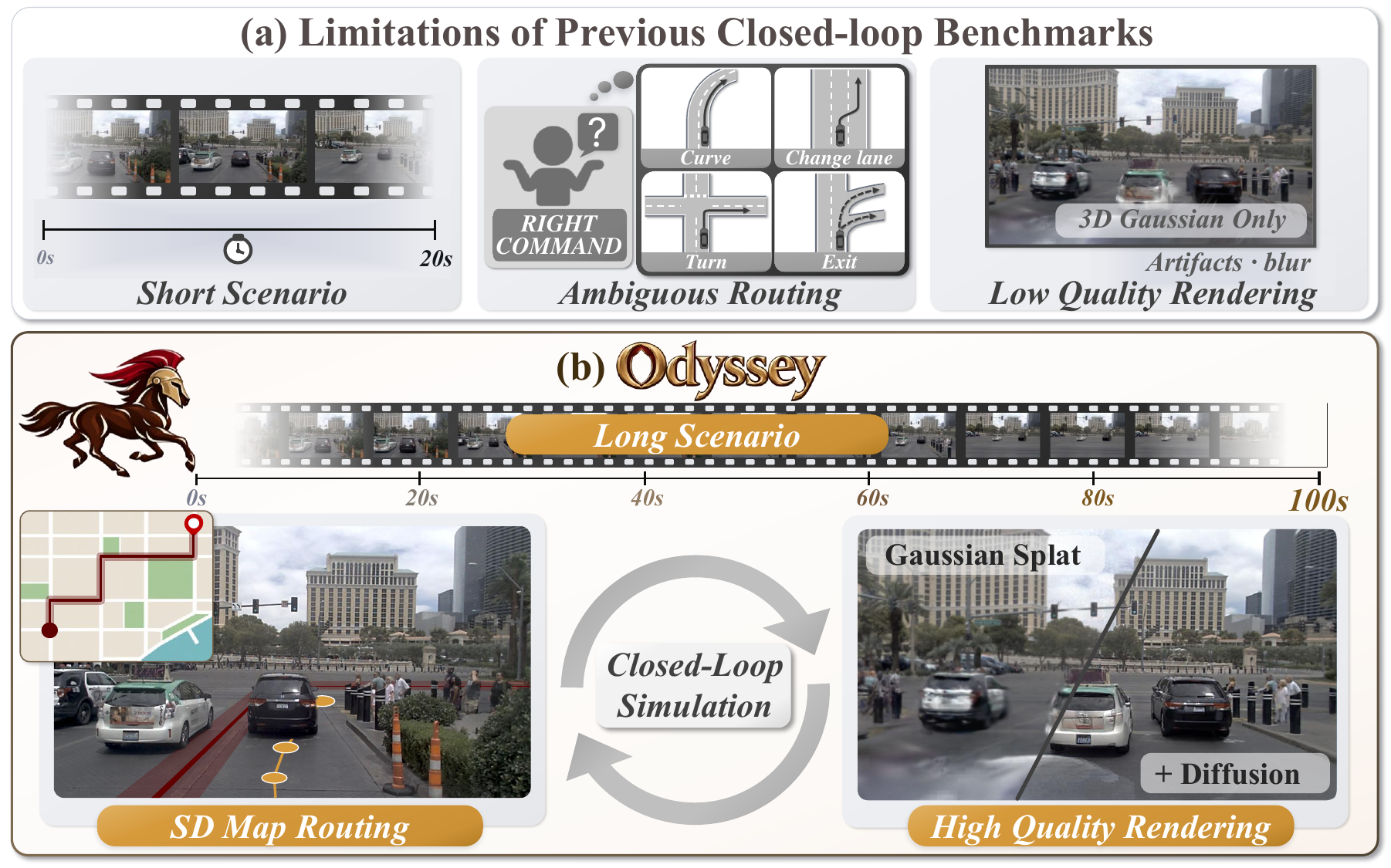}
% \caption{Overview of Odyssey, a route-aware closed-loop benchmark for real-world driving. Multi-traversal Gaussian Splatting reconstructs a geometry-consistent scene from repeated drives; reconstruction-conditioned diffusion completes partially observed vehicles and refines residual artifacts. An ego-updated OpenStreetMap route graph supplies navigation intent throughout closed-loop rollouts.}
% \caption{\textbf{Odyssey framework.} (a) Previous closed-loop benchmarks suffer from limited scenario horizons, ambiguous command-based routing, and artifacts from 3DGS-only rendering. (b) \textcolor{red}{Odyssey provides an explicit SD route and repeatedly feeds 3DGS-and-diffusion-rendered observations to the E2E model, facilitating long-horizon closed-loop rollouts, SD Map routing, and high-fidelity simulation.}}
\caption{\textbf{Odyssey framework.} (a) Previous closed-loop benchmarks suffer from limited scenario
  horizons, ambiguous command-based routing, and artifacts from 3DGS-only rendering. (b) Odyssey
  evaluates E2E planners over continuous closed-loop rollouts based on 100-second driving sequences with explicit SD-route
  guidance and diffusion-refined 3DGS observations, assessing sustained safe navigation toward the
  intended destination.}

\label{fig:intro_1}
\end{figure}

\begin{abstract}

Closed-loop evaluation of end-to-end driving requires continuous
  rollouts that reveal how earlier decisions affect subsequent driving.
  However, existing benchmarks evaluate
  only short segments and fail to capture later consequences. Ambiguous directional commands also obscure
  the intended navigation objective.
  We introduce Odyssey, a closed-loop benchmark for long-horizon
  driving comprising 100 scenarios, each reconstructed from a
  100-second nuPlan driving log to preserve the context of navigation
  maneuvers and traffic interactions.
  To provide a consistent navigation objective, Odyssey replaces
  directional commands with explicit standard-definition (SD) map
  routes that specify which roads to follow, while sensor-based
  planning determines local driving actions.
  Throughout these rollouts, diffusion-based refinement of
  3DGS-rendered images reduces rendering artifacts along the ego
  trajectory.
  To assess how effectively planners follow these routes and prepare
  for upcoming maneuvers, we introduce SD Route Compliance and
  Pre-Lane Change Score.
  These assessments are complemented by \textit{RouteDS}, which
  extends the Driving Score with penalties for SD-route deviations
  and failed lane preparation.
  We adapt state-of-the-art planners, including
  vision-language-action (VLA) models, and evaluate their
  navigation performance using these metrics.
  Odyssey highlights open questions in route representation and integration for E2E driving.
  Benchmark code and adapted baselines will be released publicly.

  \end{abstract}

\section{Introduction}

% \begin{wrapfigure}{r}{0.47\textwidth}
% \centering
% \vspace{-6pt}
% \includegraphics[width=\linewidth]{figure/intro2.pdf}
% % \caption{\textbf{One command, multiple intention.} Existing real-world evaluation setups either lack realistic sensor feedback after ego motion or provide only coarse navigation commands.}
% \caption{\textbf{Ambiguity of current command E2E system.} The same \texttt{RIGHT} command may represent distinct navigation intents: (a) continuing along a curved road, (b) making a right turn at an intersection, (c) changing to an adjacent lane, or (d) selecting one of two consecutive right exits.}
% \label{fig:motivation}
% \vspace{-10pt}
% \end{wrapfigure}

\begin{wrapfigure}{r}{0.47\textwidth}
    \vspace{-10pt}
    \centering
    \includegraphics[width=\linewidth]{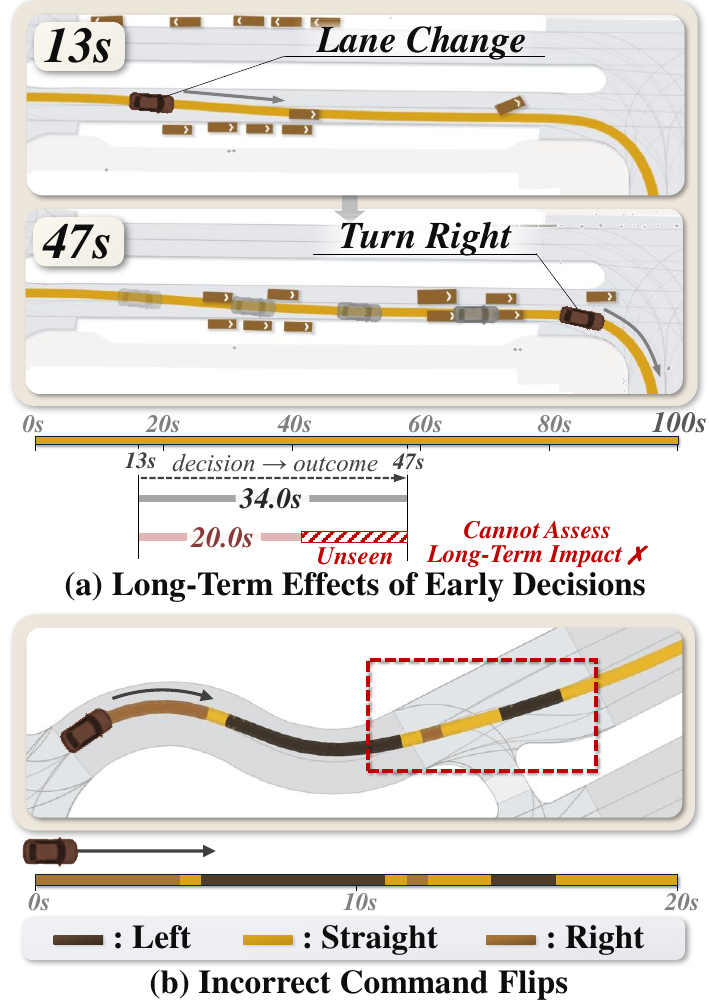}
    \vspace{-18pt}
    % \caption{\textbf{Limitations of command-based guidance.} (a) The same command can indicate different maneuvers, leaving navigation intent ambiguous. (b) Extracted commands can flip inconsistently along a continuous trajectory.}
    % \caption{\textbf{Limitations of existing evaluation setups.} (a) Short rollouts cannot assess whether an early lane change enables a later right turn. (b) Inconsistent command flips obscure navigation intent along a continuous trajectory.}
    \caption{\textbf{Limitations of existing evaluation.} (a) Short rollouts cannot assess whether a prior lane change enables a later right turn. (b) Inconsistent command flips obscure navigation intent along a continuous trajectory.}
    \label{fig:intro_2}
    \vspace{-10pt}
\end{wrapfigure}

A central challenge in autonomous driving is reliable navigation through complex traffic environments that ensure safety and comfort, as well as efficient progress toward an intended destination. 
With the growing availability of large-scale driving data, planning-oriented end-to-end (E2E) paradigms have gained prominence~\citep{uniad, diffusiondrive, safedrive, drivor}, in which driving systems directly map onboard sensor observations to driving decisions. Such planners should be evaluated by their ability to sustain safe driving as the vehicle interacts with its environment over time. Since open-loop evaluation independently compares decisions at individual time steps against recorded driving logs, it cannot capture their downstream effects. Accounting for downstream effects via closed-loop evaluation is essential to assess driving performance under evolving traffic conditions.

% 두번째 문장 Neural Rendering 주어를 뒤로 빼보자.  아니면 첫번째 문장 두번째 문장 사이에 연결어를 추가하거나
For E2E planners, closed-loop evaluation requires sensor observations that reflect the consequences of the ego vehicle's actions. Neural rendering provides this feedback by reconstructing 3D real-world scenes from driving logs and rendering new observations from the simulated ego pose. Early approaches relied on neural radiance fields (NeRFs)~\citep{unisim, neurad, emernerf}, whose volumetric rendering incurs substantial computational cost. To reduce this cost, recent methods adopt 3D Gaussian Splatting (3DGS)~\citep{drivinggaussian, splatad, hugs, omnire} for  faster rendering while maintaining high visual fidelity. These developments have supported growing research on neural-rendering-based closed-loop benchmarks~\citep{neuroncap, hugsim, alpasim} for E2E planners. However, existing benchmarks are limited in assessing a planner's ability to handle long-term driving tasks (see Figure~\ref{fig:intro_1} (a)).

% A fundamental limitation of E2E benchmarks based on real-world driving data~\citep{nuscenes, navsim, navsimv2, neuroncap, hugsim, wod_e2e} is their sole reliance on coarse left, straight, or right commands for navigation. These commands do not uniquely specify the intended route. As illustrated in Figure~\ref{fig:intro_2} (a), the same command can imply different maneuvers depending on the road context, leaving its intended meaning ambiguous. This ambiguity is particularly pronounced at two consecutive exits, where the same directional command does not uniquely specify the intended branch. Beyond this ambiguity, command extraction can produce inconsistent flips along a continuous logged trajectory (see Figure~\ref{fig:intro_2} (b)), further obscuring whether route deviations reflect planning failures or inadequate guidance. Even with clear guidance, scenarios in existing neural-rendering-based evaluation setups~\citep{neuroncap, hugsim, alpasim, instantnurec} are often reconstructed from short logs of approximately 20\,s. Such short horizons limit the assessment of how earlier decisions affect later maneuvers and whether the planner can recover.
% For example, a safe lane change to the left may cause the vehicle to miss a later right exit.
% Evaluation ending before the exit cannot reveal whether the planner moves into the exit lane in time.

One limitation of existing closed-loop benchmarks~\citep{neuroncap, hugsim} is their short evaluation horizon, typically limited to approximately 20 seconds. Such short rollouts are insufficient to fully assess the navigation capabilities of E2E planners, particularly when navigation requires a sequence of coordinated maneuvers. For example, as shown in Figure~\ref{fig:intro_2} (a), completing an upcoming right turn may require changing lanes in advance. However, the evaluation may terminate before the turn occurs, making it hard to assess the planner's ability to complete the maneuver sequence.
Furthermore, high-level meta-action commands used in many E2E benchmarks~\citep{nuscenes, navsim, navsimv2, hugsim, wod_e2e} provide ambiguous guidance for navigation in real-world driving scenarios. As illustrated in  Figure~\ref{fig:intro_1} (a), a directional command such as ``turn right'' may correspond to multiple possible maneuvers. For example, if two right exits occur in succession, the command alone does not indicate which exit the planner should take.
% \textcolor{purple}{Moreover, annotating such commands in real-world driving datasets is challenging because the driver’s intended maneuver and the appropriate timing of the command cannot always be reliably inferred from recorded trajectories} (see Figure~\ref{fig:intro_2} (b)). 
Annotating these commands in real-world driving datasets  may not accurately reflect the driver's intended maneuver~\citep{openscene}, as shown in Figure~\ref{fig:intro_2} (b).  Reliable long-horizon evaluation therefore requires more explicit navigation guidance than high-level meta-action commands alone can provide.
Another challenge is maintaining high rendering quality throughout long-horizon closed-loop simulations. When the ego vehicle deviates from the trajectory recorded during data collection, rendering quality may deteriorate because novel viewpoints lack sufficient observational support. The resulting artifacts can distort the visual input to E2E planners, influence their subsequent decisions, and undermine the reliability of both planning and evaluation. In addition, existing evaluation setups do not accurately model changes in traffic signals or pedestrian motion, further limiting their ability to represent realistic driving scenarios.

In this paper, we introduce \textbf{Odyssey}, a real-world closed-loop benchmark for evaluating E2E driving models over long-horizon rollouts. Odyssey uses neural rendering to reconstruct 3D scenes and support realistic closed-loop simulations. Each scenario includes an explicit navigation route represented on a standard-definition (SD) map. By combining long-horizon evaluation with explicit route guidance, Odyssey enables a comprehensive assessment of models’ ability to navigate complex road structures safely—a capability that existing benchmarks have not adequately evaluated.

For these extended rollouts, Odyssey builds on the explicit geometric representation of 3D Gaussian Splatting~\citep{3dgaussian}, followed by diffusion-based refinement for improving visual fidelity. Using OmniRe~\citep{omnire}, we reconstruct continuous 100-second driving environments with rigid vehicles and deformable pedestrians whose motion is controlled independently of the static background. We further apply its time-conditioned deformation field to traffic lights to model changes in their color states throughout the sequence. 
To reduce blur and recover missing detail at novel ego poses, we apply single-step diffusion refinement~\citep{difix3d, fixer} to the rendered camera images without modifying the underlying scene representation or simulated state.

% SD 라우팅에 대한 설명: SD 라우트를 제공함으로써, 100초 동안 가아할 가이던스 제시, 일관된 evaluation이 가능. 가이던스는 차의 trip 데이터 활용함, 자율주행 reasoning 능력  
% Navigation routes provide explicit guidance throughout these extended rollouts. 
 Odyssey provides navigation routes as explicit guidance  for E2E planners. 
By specifying the intended sequence of road segments, they resolve the ambiguity inherent in directional commands and eliminate the need to determine when to issue individual high-level meta-action commands. 
For each scenario, we construct an SD-map route from trip data of the data-collection fleet.
  % For each scenario, a navigation route \textcolor{purple}{is provided} based on trip data collected by the data-collection fleet. 
  However, this road-level representation encodes only coarse road-level topology and does not specify lane-level structure or the precise goal location that a driving model should reach. E2E planners must therefore interpret the high-level route jointly with the scene perceived by onboard sensors and translate it into a concrete, scene-consistent navigation goal. 
  This requires the planner to reason about how an abstract road-level route is translated to appropriate lane-level maneuvers within the current road geometry.
  Odyssey establishes a foundation for evaluating such reasoning capability of E2E planners. 
%We also provide the corresponding navigation map routes for the training set to support the development of such route-conditioned planners.  

% 어떻게 큐레이션이나 시나리오를 만들었는가, 벤치마크 구성이나 메트릭에 대해서 언급 
We design the closed-loop benchmark to enable systematic comparison of different E2E planners. We curate 100 scenarios from the raw nuPlan data~\citep{nuplan2}, covering sequences of complex navigation maneuvers and diverse road topologies. 
% \textcolor{purple}{These scenarios preserve the temporal context between preparatory lane changes and subsequent turns, enabling evaluation of whether earlier lane choices support successful navigation along the intended route.}
  % We further introduce an evaluation metric, termed \textit{RouteDS}, which penalizes deviations from the designated SD-map route, as well as collisions, lane departures, and red-light violations. 
We further introduce two evaluation metrics, termed
  \textit{Pre-Lane Change Score} and \textit{RouteDS},
  which assess whether the vehicle is appropriately positioned
  for upcoming navigation maneuvers, as well as driving safety
  and progress along the intended route.
  We evaluate a set of baseline planners adapted from recent state-of-the-art E2E and vision-language-action (VLA) models, incorporating navigation routes as additional input.
  %We believe that this formulation opens an important new research direction for autonomous driving, connecting navigation-level route planning with scene-aware, lane-level decision making in E2E driving models.   
  % These adaptations raise the broader question of how navigation
  % routes should be represented and integrated into E2E driving stacks.
  % We hope our benchmark and baselines will advance research
  % on this question.

\vspace{-1pt}
Our main contributions are as follows:
\vspace{-2pt}
\begin{itemize}
\item We introduce \textbf{Odyssey}, a high-fidelity real-world closed-loop benchmark for evaluating whether E2E planners can sustain safe and goal-directed driving under explicit route guidance over long-horizon rollouts.
\vspace{-2pt}
\item We evaluate planners over continuous rollouts based on
  recorded 100-second sequences, with explicit SD-map routes
  that resolve ambiguities in high-level driving commands
  and enable consistent assessment of navigation performance
  across complex road topologies.

\vspace{-2pt}

\item We develop a closed-loop simulation framework that combines the structural consistency of 3DGS with diffusion-based refinement to improve rendering fidelity when ego-vehicle motion reveals poorly reconstructed or previously unobserved regions.
\vspace{-2pt}
\item We will release the benchmark, evaluation code, and baseline results to enable reproducible comparison and facilitate further research on long-horizon closed-loop evaluation for autonomous driving.
\end{itemize}

\vspace{-7pt}

  \begin{table*}[t]
      \centering
      \caption{
          Comparison of driving benchmarks and simulation frameworks.
          \ensuremath{\checkmark}, \ensuremath{\triangle},
          \ensuremath{\times}, and \texttt{-} denote support,
          partial support, no documented support, and not applicable,
          respectively.
      }
      \label{tab:benchmark_comparison}

      \setlength{\tabcolsep}{3.5pt}
      \renewcommand{\arraystretch}{1.1}
      \footnotesize

      \resizebox{1.0\textwidth}{!}{%
      \begin{tabular}{@{}lcccccccc@{}}
          \toprule[1pt]
          \makecell[l]{\textbf{Benchmark}}
          & \makecell{\textbf{Real-log}\\\textbf{scenarios}}
          & \makecell{\textbf{Sensor}\\\textbf{closed-loop}}
          & \makecell{\textbf{Realistic}\\\textbf{images}}
          & \makecell{\textbf{Navigation}\\\textbf{input}}
          % & \makecell{\textbf{Scenario}\\\textbf{horizon}}
          & \makecell{\textbf{Recorded sequence}\\\textbf{duration}}
          & \makecell{\textbf{Rendering}\\\textbf{type}}
          & \makecell{\textbf{Traffic-light}\\\textbf{state}}
          & \makecell{\textbf{Pedestrian}\\\textbf{motion}}
          \\
          \specialrule{\lightrulewidth}{0.25ex}{0.50ex}

          nuScenes
          & $\checkmark$ & $\times$ & $\checkmark$
          & Commands
          & - & $\times$
          & - & - \\

          NAVSIM
          & $\checkmark$ & $\times$ & $\checkmark$
          & Commands
          & - & $\times$
          & - & - \\

          NAVSIMv2
          & $\checkmark$ & $\times$ & $\checkmark$
          & Commands
          & - & Offline 3DGS
          & - & - \\

          WOD-E2E
          & $\checkmark$ & $\times$ & $\checkmark$
          & Commands
          & - & $\times$
          & - & - \\

          \specialrule{\lightrulewidth}{0.25ex}{0.50ex}

          nuPlan
          & $\checkmark$ & $\times$ & $\times$
          & HD route
          & $15\,\mathrm{s}$ & $\times$
          & - & - \\

          Waymax
          & $\checkmark$ & $\times$ & $\times$
          & HD route
          & $9\,\mathrm{s}$ & $\times$
          & - & - \\

          \specialrule{\lightrulewidth}{0.25ex}{0.50ex}

          MetaDrive
          & $\triangle$ & $\checkmark$ & $\times$
          & Target points
          & User-set & Rasterization
          & $\checkmark$ & $\checkmark$ \\

          CARLA Leaderboard
          & $\times$ & $\checkmark$ & $\times$
          & Target points
          & $\sim3$--$5\,\mathrm{min}$ & Rasterization
          & $\checkmark$ & $\checkmark$ \\

          Bench2Drive
          & $\times$ & $\checkmark$ & $\times$
          & Target points
          & $\sim15$--$30\,\mathrm{s}$ & Rasterization
          & $\checkmark$ & $\checkmark$ \\

          \specialrule{\lightrulewidth}{0.25ex}{0.50ex}

          NeuroNCAP
          & $\checkmark$ & $\checkmark$ & $\checkmark$
          & Commands
          & $20\,\mathrm{s}$ & NeRF
          & $\times$ & $\times$ \\

          HUGSIM
          & $\checkmark$ & $\checkmark$ & $\checkmark$
          & Commands
          & $\leq20\,\mathrm{s}$ & 3DGS
          & $\times$ & $\times$ \\

          DriveArena
          & $\triangle$ & $\checkmark$ & $\checkmark$
          & Commands
          & $20\,\mathrm{s}$ & Diffusion
          & $\times$ & $\times$ \\

          \specialrule{\lightrulewidth}{0.25ex}{0.50ex}

          \textbf{Odyssey (ours)}
          & $\boldsymbol{\checkmark}$
          & $\boldsymbol{\checkmark}$
          & $\boldsymbol{\checkmark}$
          & \textbf{SD route}
          & $\mathbf{100}\,\mathbf{s}$
          & \textbf{3DGS + diffusion}
          & $\boldsymbol{\checkmark}$
          & $\boldsymbol{\checkmark}$ \\

          \bottomrule[1pt]
      \end{tabular}%
      }
  \end{table*}
\vspace{-5pt}

% \begin{figure}[t]
% \centering
% \includegraphics[width=\columnwidth]{figure/Method1.pdf}
% % \caption{Overview of Odyssey, a route-aware closed-loop benchmark for real-world driving. Multi-traversal Gaussian Splatting reconstructs a geometry-consistent scene from repeated drives; reconstruction-conditioned diffusion completes partially observed vehicles and refines residual artifacts. An ego-updated OpenStreetMap route graph supplies navigation intent throughout closed-loop rollouts.}
% \caption{\textbf{Odyssey Overall Architecture.} I am painting the overall method.}
% \label{fig:overview}
% \end{figure}

\section{Related Work}
Open-loop displacement metrics overlook how planning actions
  affect subsequent driving~\citep{nuscenes, stp3, uniad}.
  NAVSIM introduces simulation-based driving metrics~\citep{navsim},
  and NAVSIM v2 adds pre-rendered feedback~\citep{navsimv2}.
  Continuous replanning further requires observations from
  the ego poses reached after executing successive actions.
  Neural simulators provide this feedback in reconstructed
  environments~\citep{unisim, neurad, neuroncap, hugsim, alpasim}.
  However, short evaluation sequences can end before earlier
  lane choices affect later navigation~\citep{neuroncap, hugsim}.

  Assessing these consequences also requires a consistent
  navigation objective.
  Directional commands can leave consecutive branches
  ambiguous~\citep{uniad, vad, navsim, hugsim}, obscuring whether
  route deviations reflect planning errors or insufficient guidance.
  Nearby target points provide spatial guidance but limited
  context for upcoming route choices~\citep{carla, bench2drive}.
  Odyssey addresses these temporal and navigational limitations
  through extended closed-loop rollouts with explicit SD-map
  routes, assessing whether successive decisions support
  safe progress along the intended road sequence.

\section{Odyssey}
\label{sec:odyssey}
Odyssey provides a closed-loop simulation platform for evaluating long-horizon navigation along explicit routes on an SD map. Within each scene reconstructed from a 100-second driving log, the planner predicts a local driving trajectory based on camera observations and route guidance. The simulator advances the ego vehicle along the predicted trajectory and uses a neural rendering engine to generate new camera observations from its updated pose. We evaluate planner performance in terms of driving safety, route progress, adherence to the road-level route, and lane selection before intersections.
\vspace{-5pt}

\subsection{Scenario Curation }
  \label{sec:long_horizon}
\begin{wrapfigure}{r}{0.44\textwidth}
      \centering
      \vspace{-10pt}
      \includegraphics[width=\linewidth]{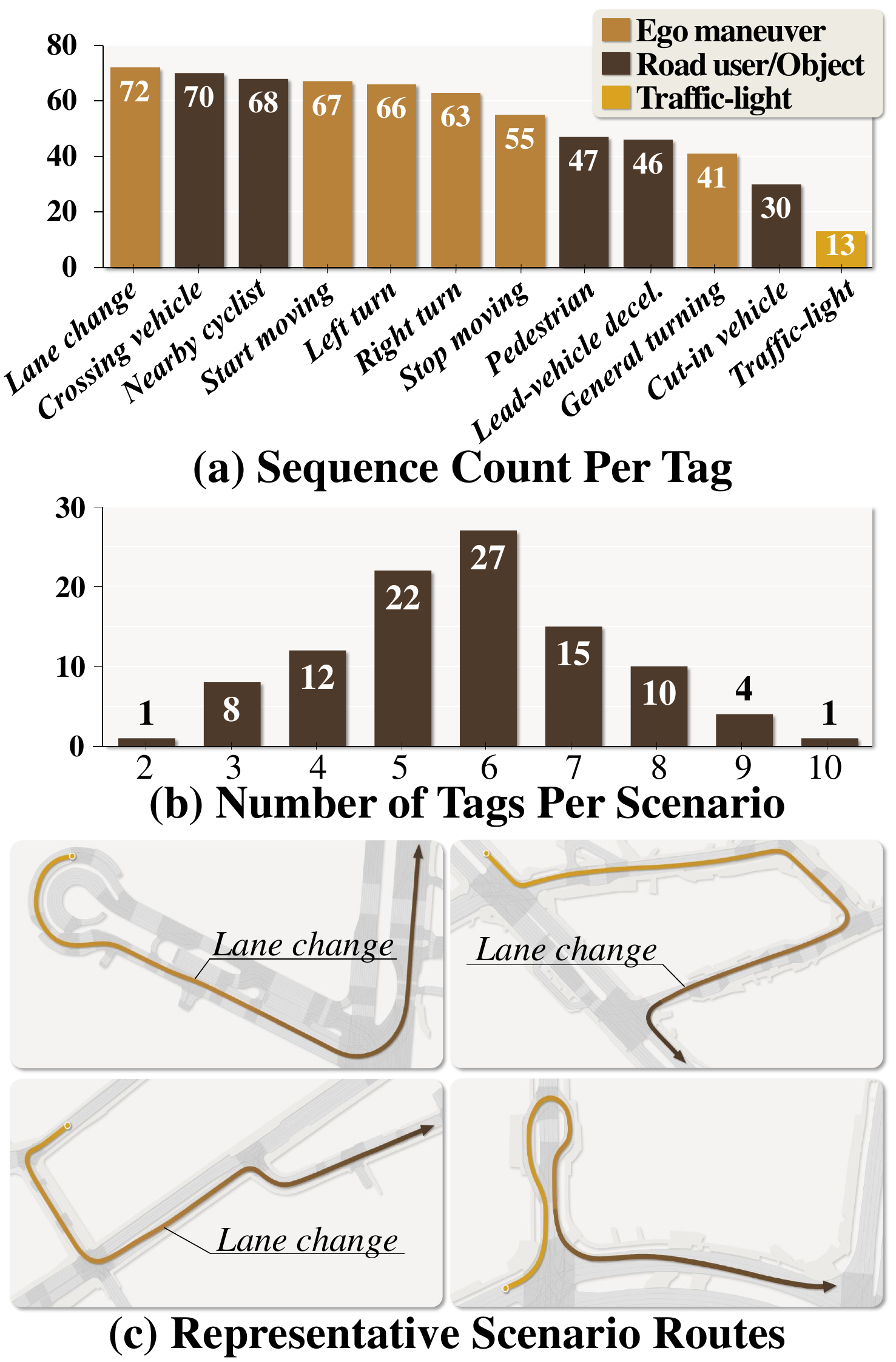}
      % \vspace{-15pt}
      % \caption{\textbf{Scenario-tag distributions in Odyssey.}
      % (a) Distribution of the number of tags per scenario.
      % (b) Scenario counts and percentages for each tag.
      % Tags are not mutually exclusive.}
      \caption{Scenario composition and example routes in Odyssey.
  % (a) Distribution of the number of tags per scenario.
  % (b) Number of scenarios containing each tag, grouped into
  % ego maneuvers, road users and objects, and traffic lights.
  % Tags are not mutually exclusive.
  % (c) Representative scenario routes across different road layouts,
  % with lane-change locations annotated.
  }
      \label{fig:curation}
      \vspace{-70pt}
  \end{wrapfigure}

Odyssey comprises 100 driving scenarios curated from
nuPlan~\citep{nuplan2}.
To capture diverse driving situations over extended sequences,
we combine RefAV~\citep{refav}-based tagging with human review
of navigation maneuvers and traffic interactions.
The distribution of individual tags is reported in
Figure~\ref{fig:curation} (a).
The selected scenarios contain multiple maneuver and interaction
tags, with the number of tags per scenario shown in
Figure~\ref{fig:curation} (b).
The selected set includes 13 scenarios involving traffic lights
and 47 involving pedestrians, covering both signal compliance
and safety around vulnerable road users.

  Within this set, we prioritize scenarios with complex road layouts and sequences of navigation maneuvers. Figure~\ref{fig:curation} (c) shows representative routes involving successive turns, branching roads, and lane changes. 
In particular, scenarios with a lane change followed by a turn allow us 
 to assess whether the planner changes lanes in preparation for the turn. 
  The set includes 49 such scenarios, 16 of which have more than 20 seconds between the start of the lane change and the start of the turn. A single 20-second evaluation window cannot capture the onset of both maneuvers in these cases. 
  
  %By preserving this temporal context in continuous 100-second rollouts, Odyssey enables evaluation of whether earlier lane choices support subsequent turns along the intended route.

% We identify these situations through RefAV-based~\citep{refav} and human tagging of the driving logs. We select 100 second sequences to reduce repeated coverage of the same locations, while retaining a subset of scenes with fewer tagged interactions. The resulting automatic selection, before later human boundary edits, contains 141 source sequences, including 63 representative lane-change--turn sequences and 30 pedestrian-related sequences. In 21 of the 63 lane-change \& turn sequences, the lane change and turn begin more than 20 s apart; a single 20 s window therefore cannot contain both events.

% \begin{figure}[h]
% \centering
% \includegraphics[width=1.0\columnwidth]{figure/Curation}
% \vspace{-10pt}
% \caption{\textbf{Scenario-tag distributions in Odyssey.}
%       (a) Distribution of the number of tags per scenario.
%       (b) Scenario counts and percentages for each tag.}
% \label{fig:curation}
% \end{figure}

\vspace{5pt}
\subsection{SD Map Routing}
\vspace{5pt}
  \label{sec:sd_routing}
SD maps represent the road network through coarse polylines and connectivity at intersections and branches. A navigation route specifies the sequence of roads to follow. To construct each route, we match the full logged ego trajectory to OpenStreetMap (OSM) using HMM-based map matching~\citep{newson2009hidden}, yielding a single polyline from the scenario start to the destination. We manually check the matched roads and route continuity against the logged trajectory and SD map. During closed-loop simulation, we update the route segment provided to the planner according to the ego vehicle’s current position.

\vspace{5pt}
\subsection{Closed-Loop Simulations}
\vspace{5pt}
\label{sec:simulator}

\noindent\textbf{\scshape 3D Gaussian Scene Reconstruction.} 
Odyssey reconstructs each 3D sequence using
  OmniRe~\citep{omnire}, with rigid transformations for vehicles,
  SMPL-based deformation~\citep{loper2015smpl} for pedestrians,
  and a learned deformation field for other actors.
  Traffic lights have fixed geometry and position, with
  time-dependent spherical-harmonic coefficients and opacity
  capturing signal changes.
  To stabilize road geometry, we incorporate
  \textit{ground modeling} inspired by HUGSIM~\citep{hugsim},
  representing the road surface separately using thin Gaussian
  disks initialized from LiDAR.
  Their heights and normals follow a fixed LiDAR-derived height
  field as we optimize their in-plane geometry and appearance.
  Opacity regularization suppresses overlapping background
  Gaussians near the road surface.

  When training these scene representations over long sequences,
  uniform image sampling concentrates updates in regions
  where the recording vehicle stops or moves slowly.
  We therefore apply \textit{coverage-aware sampling},
  weighting each view by aggregating inverse observation counts
  over its visible scene voxels.
  This increases sampling probabilities for views covering
  less-observed regions.

\noindent\textbf{\scshape Reconstruction-Conditioned Visual Refinement.}
Rendering from ego poses beyond the logged viewpoints can
  expose blur and missing detail in poorly observed regions.
  We reduce these artifacts using Fixer~\citep{fixer},
  which adapts the diffusion-based refinement of
  Difix3D+~\citep{difix3d} to a Cosmos backbone~\citep{cosmos_predict2}.
  Fixer refines each image in a single denoising step
  at a fixed timestep without adding random noise.
  We fine-tune it offline on paired degraded renderings
  and recorded images, following the data preparation
  strategy of Difix.
  During simulation, the model refines rendered observations
  before they are passed to the planner.

% Scene reconstruction is supervised by viewpoints available in the driving logs, leaving poorly observed regions susceptible to blur and missing detail when rendered from novel ego poses. To reduce these artifacts, we use Fixer~\citep{fixer}, which adapts the diffusion-based refinement approach of Difix~\citep{difix3d} to a Cosmos backbone~\citep{cosmos_predict2}. Fixer processes each rendered image in a single denoising step at a fixed timestep without injecting random noise, avoiding sampling-induced variation in the visual feedback for a given input. 
% To adapt Fixer to Odyssey, we fine-tune it offline on paired   degraded renderings and corresponding recorded images, following the data preparation strategy of Difix. During simulation, the fine-tuned model refines rendered images from the current ego viewpoint without modifying the underlying Gaussian representation or simulated state. 

% The ego vehicle follows the planner's predicted trajectory using a linear quadratic regulator (LQR)~\citep{anderson2007optimal} for tracking control and a bicycle model~\citep{kong2015kinematic} to update its state. The LQR computes steering and acceleration commands by balancing tracking error against control effort. The bicycle model updates the ego state according to these commands, relating heading changes to vehicle speed, steering, and wheelbase. Each planner prediction updates the tracking reference, while execution proceeds from the current simulated state. 
\noindent\textbf{\scshape Ego Vehicle Control.}
% The ego vehicle follows the planner's predicted trajectory
%   through a control and state-update loop running at 10\,Hz.
%   At each step, a linear quadratic regulator
%   (LQR)~\citep{anderson2007optimal} computes steering and acceleration
%   commands by balancing tracking error against control effort.
%   These commands are then applied to a bicycle
%   model~\citep{kong2015kinematic} to update the ego state,
%   relating heading changes to vehicle speed, steering, and wheelbase.
The ego vehicle tracks the planner's trajectory
  at 10\,Hz using a linear quadratic regulator
  (LQR)~\citep{anderson2007optimal}.
  At each step, the LQR computes acceleration and steering rate that a bicycle model~\citep{kong2015kinematic}
  uses to update the ego state.

\begin{figure}[t]
\centering
\includegraphics[width=1.0\columnwidth]{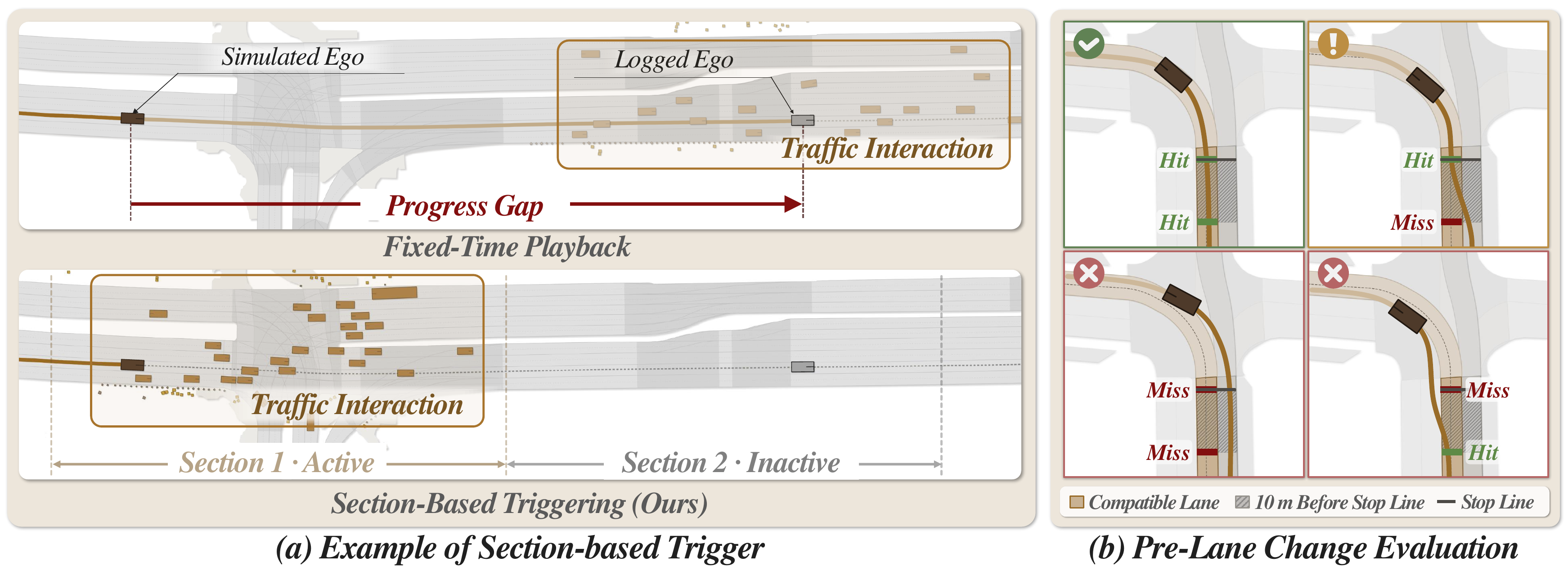}
\vspace{-15pt}
\caption{\textbf{Traffic triggering and pre-lane change evaluation.}
  (a) When the simulated ego is slower than the logged ego,
  fixed-time playback (top) places traffic interactions ahead of it.
  By activating traffic as the ego approaches a section and
  keeping subsequent sections inactive, section-based
  triggering (bottom) aligns traffic interactions with ego progress.
  (b) Pre-Lane Change requires both driving in route-compatible
  lanes throughout the final $10\,\mathrm{m}$ and crossing
  the stop line in a compatible lane.
  The four examples illustrate cases that satisfy both
  conditions, only one, or neither.
}
  \label{fig:traffic_lane}
\end{figure}

\subsection{Traffic Simulation}
  \label{sec:traffic_simulation}

 \noindent\textbf{\scshape Traffic Configuration.}
 Following nuPlan~\citep{nuplan1}, we evaluate planners under
  non-reactive traffic, where other actors replay
  recorded motion, and reactive traffic, where surrounding
  vehicles use IDM-based longitudinal control~\citep{idm}.
  Vehicles enter the simulation as they appear in the logs.
  At unprotected intersections, newly introduced vehicles
  can create conflicts that longitudinal control alone
  cannot resolve.
  We therefore add intersection-handling logic inspired by
  CARLA's Traffic Manager~\citep{carla} to prevent deadlocks.

% We evaluate planners under non-reactive and reactive traffic settings, following nuPlan~\citep{nuplan1}. In both settings, the evaluated planner controls the ego vehicle. Surrounding actors replay their recorded motion in the non-reactive setting, and surrounding vehicles respond to simulated traffic through IDM-based longitudinal control~\citep{idm} in the reactive setting. Our extended rollouts additionally require vehicles to enter the simulation as they appear in the logs. During unprotected intersection maneuvers, newly introduced vehicles can create conflicting interactions that longitudinal control alone cannot resolve, leaving vehicles stuck. We therefore augment the reactive controller with intersection-handling logic inspired by CARLA's Traffic Manager~\citep{carla}.
% \vspace{-10pt}

  \noindent\textbf{\scshape Section-Based Traffic Triggering.}
    When the ego vehicle travels faster or slower than the logged
  vehicle, traffic playback on the original timeline can cause
  it to miss recorded interactions.
  We therefore introduce section-based traffic triggering
  to align traffic activation and traffic light signal timing
  with ego progress in both non-reactive and reactive settings.
  Since traffic interactions are concentrated around intersections,
  we place section boundaries between intersections.
  As the ego vehicle approaches a section, we activate its traffic
  using a fixed time offset determined by the difference between
  the logged and simulated ego entry times.
  The same offset is applied to the section's traffic light signals,
  which then advance through their recorded states at the original
  rate.
  Figure~\ref{fig:traffic_lane} (a) compares this approach
  with traffic playback on the original timeline.

  \subsection{Evaluation Protocol}
  \label{sec:evaluation_protocol}
  We introduce \textit{SD Route Compliance},
  \textit{Pre-Lane Change Accuracy},
  \textit{Pre-Lane Change Score}, and \textit{RouteDS}
  to assess navigation performance.
  We also report Efficiency and Comfortness.

  \noindent\textbf{\scshape SD Route Compliance (SDC).}
  We map-match the executed ego trajectory to assess adherence
  to the provided SD-map route.
  A rollout scores 1 if all matched road segments belong
  to the route and 0 otherwise, even if the vehicle stays
  on drivable roads.
  SDC is the mean score across rollouts.

  \noindent\textbf{\scshape Pre-Lane Change Accuracy (PLCA) \& Score (PLCS).}
  Long-horizon navigation requires timely lane positioning
  to follow the route through successive intersections.
  We assess this ability via PLCA and PLCS at intersections
  where continuing along the route requires specific approach lanes.
  PLCA averages scores over evaluated intersections encountered
  across all rollouts.
  PLCS averages scores over all evaluation intersections
  along the full scenario route, counting unencountered
  intersections as 0 to account for incomplete rollouts.
  At each stop line, the vehicle scores 1 if it crosses
  in a route-compatible lane and has maintained compatible
  lane occupancy for the preceding 10\,m.
  A later entry scores 0.5, and crossing outside a compatible
  lane scores 0 (Fig.~\ref{fig:traffic_lane} (b)).

  \noindent\textbf{\scshape RouteDS.}
  We extend the Driving Score of Bench2Drive~\citep{bench2drive}
  with penalties for SD-route deviations and failed
  pre-lane change evaluations.
  These additions complement route completion and penalties
  for at-fault collisions~\citep{navsim}, lane deviation,
  and red-light violations.
  For each scenario, the route completion ratio is multiplied
  by a cumulative penalty factor initialized to 1,
  with multipliers in $[0,1]$ applied for infractions.
  Each failed pre-lane change evaluation incurs a separate penalty.
  We report the mean RouteDS across scenarios.
  Detailed metric definitions and penalty values
  are provided in the Appendix.

\section{Implementation Details}

\subsection{Simulator Implementation Details}

We reconstruct the entire 100-second sequence as a single scene using OmniRe~\citep{omnire} and accelerating optimization with FastGS~\citep{fastgs}. Each scene contains 8,000 images captured at 10 Hz across eight cameras and is optimized for 140,000 iterations. We additionally apply per-image affine color corrections, as used in MTGS~\citep{mtgs}, to compensate for exposure variations.
We fine-tune pretrained Fixer~\citep{fixer} on paired degraded
  renderings and corresponding recorded images, following
  the data preparation strategy of Difix~\citep{difix3d}.
  Details of training data generation and fine-tuning
  are provided in the Appendix.

\begin{table*}[!t]
  \centering
\begin{minipage}[t]{0.58\textwidth}
      \centering
      \caption{Ablation study of reconstruction components
      on Odyssey. KID$_{\mathrm{in}}$ and KID$_{\mathrm{ex}}$
      evaluate interpolated and extrapolated views, respectively.}
      \label{tab:reconstruction_quality}
      \renewcommand{\arraystretch}{1.1}
      \setlength{\tabcolsep}{5pt}

      \resizebox{\linewidth}{!}{%
      \begin{tabular}{l c c c c c c}
          \toprule[1pt]
          \textbf{Method}
          & \textbf{PSNR} $\uparrow$
          & \textbf{SSIM} $\uparrow$
          & \textbf{LPIPS} $\downarrow$
          & \textbf{KID$_{\mathrm{in}}$} $\downarrow$
          & \textbf{KID$_{\mathrm{ex}}$} $\downarrow$
          & \makecell[c]{\textbf{Latency}\\
            \textbf{(ms/image)} $\downarrow$} \\
          \midrule[0.5pt]

          OmniRe (Baseline)
          & 24.34 & 0.788 & 0.399 & 29.5 & 52.2 & \textbf{17.8} \\

          \quad + Color Correction
          & 24.98 & 0.793 & 0.392 & 25.4 & 47.0 & 18.1 \\

          \quad + Coverage-aware Sampling
          & 25.22 & 0.796 & 0.382 & 24.1 & 42.8 & 18.1 \\

          \quad + Ground Modeling
          & 25.20 & 0.794 & 0.376 & 20.1 & 32.8 & 32.4 \\

          \quad + Pretrained Fixer
          % & 25.00 & \textbf{0.801} & 0.330 & 14.6 & 17.6 & 57.4 \\ trt 안 쓴 속도
          & 25.00 & \textbf{0.801} & 0.330 & 14.6 & 17.6 & 39.1 \\

          \quad + Fixer Fine-tuning \textbf{(Ours)}
          & \textbf{25.34} & 0.786 & \textbf{0.233}
          & \textbf{5.6} & \textbf{8.9} & 39.1 \\

          \bottomrule[1pt]
      \end{tabular}%
      }
  \end{minipage}
\hfill
  % Right: visual refinement ablation
  \begin{minipage}[t]{0.40\textwidth}
      \centering
      \caption{Effect of Fixer refinement on perception and driving performance.}
      \vspace{6pt}
      \label{tab:refinement_ablation}
      \renewcommand{\arraystretch}{0.95}
      \setlength{\tabcolsep}{5pt}

      \resizebox{\linewidth}{!}{%
      \begin{tabular}{l | c | c c}
          \toprule[1pt]

          \multirow[c]{2}{*}{\textbf{Observation}}
          & \textbf{Open-loop}
          & \multicolumn{2}{c}{\textbf{Closed-loop}} \\
          \cmidrule(lr){3-4}

          & \textbf{mIoU} $\uparrow$
          & \textbf{mIoU} $\uparrow$
          & \textit{\textbf{RouteDS}} $\uparrow$ \\
          \midrule[0.5pt]

          GT Image
          & 35.4 & -- & -- \\
          \midrule[0.5pt]

          3D GS Only
          & 31.5 & 26.0 & 53.80 \\

          \quad + Diffusion
          & 31.9 & 28.2 & 52.62 \\

          % \quad + Fine-tuning \textbf{(Ours)}
          \quad + Fine-tuning
          & \textbf{33.7} & \textbf{29.4} & \textbf{54.43} \\

          \bottomrule[1pt]
      \end{tabular}%
      }
  \end{minipage}

\end{table*}

\begin{figure}[!t]
\centering
\includegraphics[width=\columnwidth]{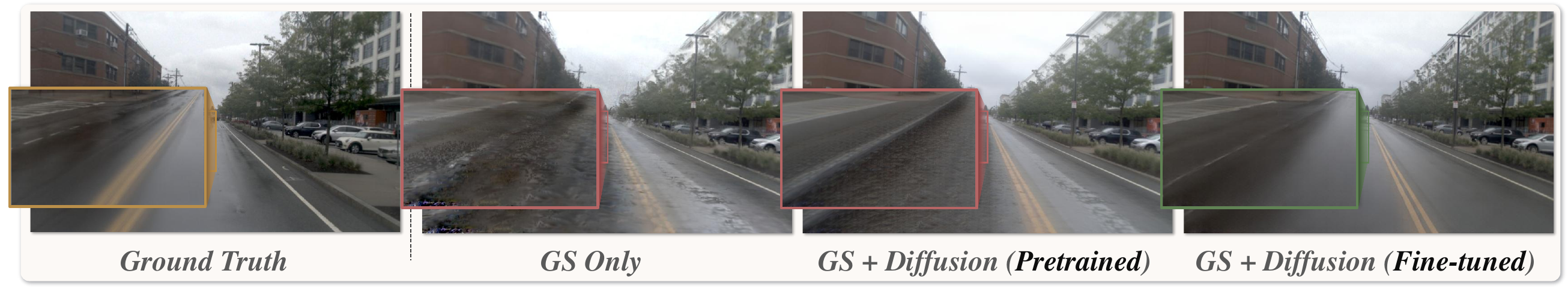}
  \caption{\textbf{Effect of diffusion refinement at a novel viewpoint.}
The first image is the recorded image, and the other three
  are rendered from a viewpoint shifted $2\,\mathrm{m}$ to the
  left of the recorded camera position.
  GS-only rendering exhibits distorted road textures at this
  novel viewpoint.
  Pretrained Fixer improves visual quality but introduces
  spurious curb-like structures on the road.
  Fine-tuning on Odyssey suppresses these artifacts and yields smoother roads and clearer lane markings.}
\vspace{-10pt}
\label{fig:gs_comparison}
\end{figure}

\subsection{SD-Route-Conditioned Baselines}
\label{main:sd_baseline}
We adapt existing planners to SD-route guidance and compare
  their closed-loop driving performance over long-horizon rollouts.
  The baselines comprise LTF~\citep{transfuser},
  DiffusionDrive~\citep{diffusiondrive},
  DrivoR~\citep{drivor}, SafeDrive~\citep{safedrive},
  and ReCogDrive~\citep{recogdrive}.
  All baselines predict a 4-second future trajectory sampled
  at 2\,Hz and are trained on the NAVSIM training set~\citep{navsim}.

  At each planning step, we sample 120 route points at
  1-meter intervals ahead of the ego's projection onto
  the SD-route polyline.
  Their positions, directions, and headings are expressed
  in the ego frame to provide navigation guidance in place
  of driving commands.
  For vision-based planners, a linear projection and max pooling
  over groups of five consecutive points produce 24 route tokens.
  Fourier embeddings of group starting poses and learned order
  embeddings encode their spatial placement and sequence.
  Each planning decoder layer attends to these tokens as keys
  and values through an additional cross-attention module
  before its feed-forward network.
  For ReCogDrive, we summarize the route into a single token
  added to the ego-status features and supply a route-derived
  navigation sentence to the VLM.
  Model-specific adaptations are detailed in the Appendix.

  Dedicated reasoning mechanisms that relate SD-map routes
  to the perceived scene and required driving maneuvers
  are left to future work.

% We select representative models evaluated on the NAVSIM benchmark to cover a diverse set of E2E planning paradigms. Specifically, we include LTF, the imitation-learning baseline provided by NAVSIM; DiffusionDrive, which employs a diffusion-based trajectory decoder; SafeDrive, which leverages instance-centric interaction and sparse world modeling; and DrivoR, which achieves strong performance on NAVSIM PDMS through safety scoring. To further support a broader range of future research on Odyssey, we also provide VLA-based E2E planners, including ReCogDrive and DriveVLA-W0, as additional baselines.

% To adapt existing planners to Odyssey's SD-route guidance, our default baseline removes the original driving-command input and encodes the SD route into a global route feature. This feature is concatenated with the ego-status feature and projected through an MLP to construct the status embedding, while minimizing changes to the original planner architectures. Additional route-conditioning strategies that inject SD-route information into the planning query or trajectory decoder are described in Appendix.
% \subsection{SD Map Routing Encoding Module}
% \subsubsection{Right \& Left Embedding}

\section{Experiments}

  \begin{wrapfigure}{r}{0.45\textwidth}
    \centering
    \vspace{-30pt}
    \includegraphics[width=\linewidth]{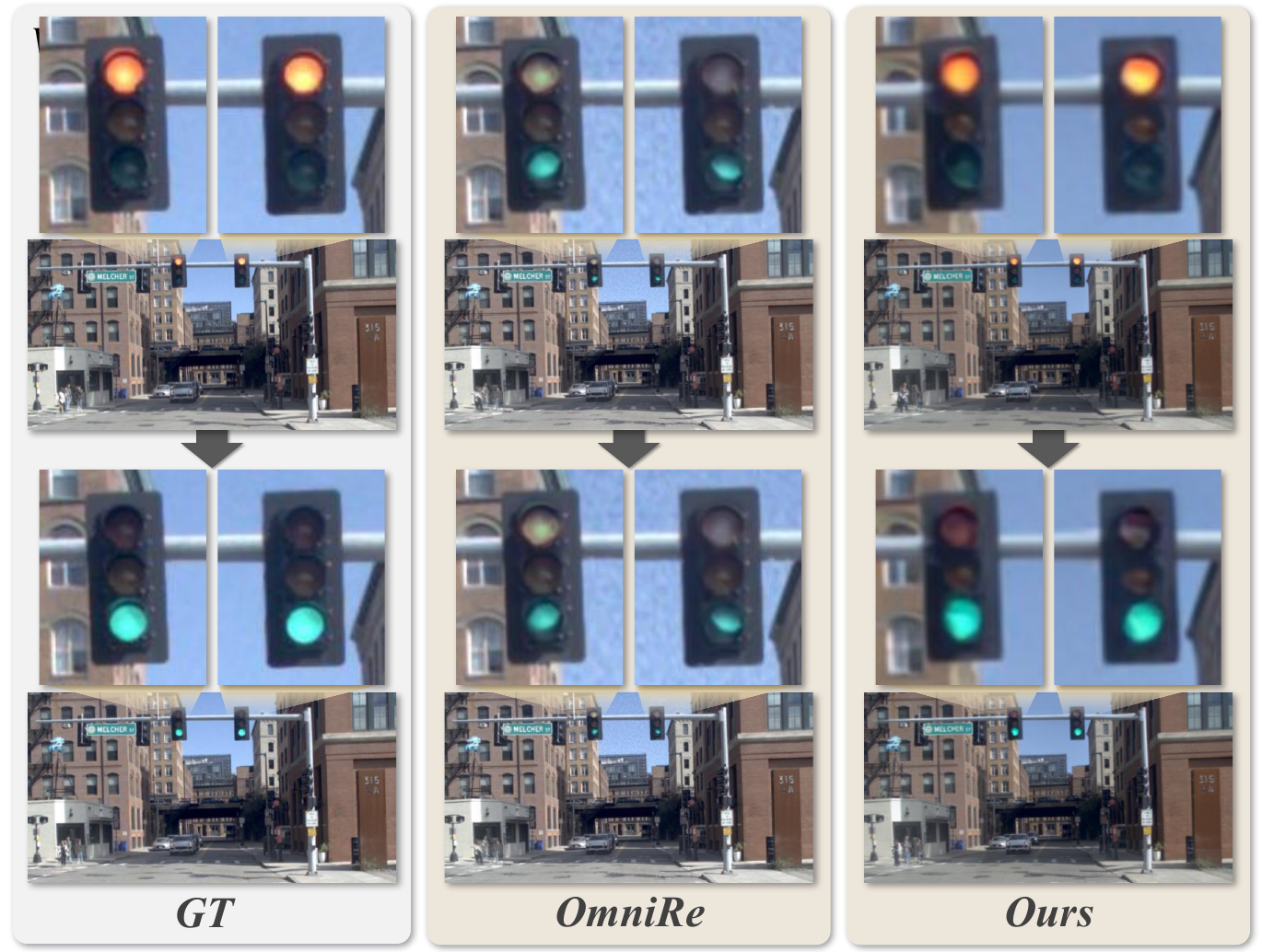}
    \vspace{-15pt}
    \caption{\textbf{Traffic-light modeling.}
    Our deformation field reproduces the red-to-green
    transition, which OmniRe fails to recover.}
    \label{fig:traffic}
    \vspace{-20pt}
  \end{wrapfigure}%

  \subsection{Scene Reconstruction Quality}
  We evaluate reconstruction quality using PSNR, SSIM, LPIPS,
  and KID~\citep{kid}.
  For novel viewpoints without paired ground truth, we follow
  HUGSIM~\citep{hugsim} and use KID to measure the squared
  maximum mean discrepancy between Inception features
  extracted from rendered and recorded images. Table~\ref{tab:reconstruction_quality} shows cumulative reductions in LPIPS and KID
  % 앞뒤에 빈 줄을 넣지 않습니다.
  as color correction, coverage-aware sampling, ground modeling, and diffusion
  refinement are incorporated.
  The visual improvements are illustrated in
  Figure~\ref{fig:gs_comparison}, where Fixer fine-tuning
  suppresses road-surface artifacts and improves lane clarity.
  Alongside image quality, the table reports per-image latency
  measured on an NVIDIA B200 GPU.

  We also examine how reconstruction captures changes in
  traffic-light appearance.
  Figure~\ref{fig:traffic} shows that our time-conditioned
  appearance model reproduces the red-to-green transition
  missed by OmniRe.
  The rendered signals provide visual cues for evaluating
  planner responses to traffic lights during closed-loop rollouts.

\begin{table*}[!t]
      \centering
      \caption{Comparison of baseline planners with and without
      SD-route guidance under non-reactive and reactive traffic.
      Eff. and Comf. denote Efficiency and Comfortness, respectively.
      Bold and underlined values indicate the best and second-best
      results in each metric, respectively.}
      \vspace{-5pt}
      \label{tab:sdroute_baselines}
      \renewcommand{\arraystretch}{0.95}
      \setlength{\tabcolsep}{3pt}

      \resizebox{\textwidth}{!}{%
      \begin{tabular}{l c | c c c c c c | c c c c c c | c}
          \toprule[1.3pt]

          \multirow[c]{2}{*}[-0.5ex]{\textbf{Method}}
          & \multirow[c]{2}{*}[-0.5ex]{%
              \shortstack[c]{\textbf{SD-Route}\\\textbf{Guidance}}}
          & \multicolumn{6}{c|}{\textbf{Odyssey Non-reactive}}
          & \multicolumn{6}{c|}{\textbf{Odyssey Reactive}}
          & \makecell[c]{\textbf{NAVSIM}\\\textbf{Open Loop}} \\

          \cmidrule(lr){3-8}
          \cmidrule(lr){9-14}
          \cmidrule(lr){15-15}

          &
          & \textit{\textbf{RouteDS}}
          & \textbf{SDC}
          & \textbf{PLCA}
          & \textbf{PLCS}
          & \textbf{Eff.}
          & \textbf{Comf.}
          & \textit{\textbf{RouteDS}}
          & \textbf{SDC}
          & \textbf{PLCA}
          & \textbf{PLCS}
          & \textbf{Eff.}
          & \textbf{Comf.}
          & \textbf{PDMS} \\
          \midrule[0.85pt]

          \multicolumn{15}{c}{\textit{Vision-based planners}} \\
          \midrule[0.3pt]

          \multirow{2}{*}{LTF}
          & $\times$
          & 16.2 & 51 & 66.0 & 50.3 & 63.9 & \textbf{100}
          & 16.3 & 54 & 63.5 & 49.1 & 76.1 & \underline{99} & 83.8 \\
          & $\checkmark$
          & 20.1 & \underline{95} & 57.1 & 56.1 & 68.2 & 96
          & 23.0 & \underline{93} & 53.6 & 52.6 & 78.0 & 93 & 85.6 \\
          \midrule[0.3pt]

          \multirow{2}{*}{DiffusionDrive}
          & $\times$
          & 25.6 & 59 & 78.8 & 63.7 & 61.4 & \textbf{100}
          & 24.2 & 59 & 77.0 & 63.1 & 75.7 & \textbf{100} & 85.8 \\
          & $\checkmark$
          & 35.9 & \textbf{100} & \textbf{82.0} & \textbf{80.5} & 63.4 & \underline{99}
          & 41.0 & \textbf{100} & \textbf{81.2} & \textbf{80.2} & 71.5 & \underline{99} & 86.9 \\
          \midrule[0.3pt]

          \multirow{2}{*}{DrivoR}
          & $\times$
          & 20.6 & 50 & 75.5 & 62.8 & \textbf{104.7} & 44
          & 25.5 & 53 & 76.6 & 62.8 & \textbf{133.4} & 47 & \textbf{93.1} \\
          & $\checkmark$
          & \underline{36.5} & 90 & 72.8 & 66.0 & \underline{84.5} & 60
          & \underline{41.4} & 89 & 71.8 & 66.0 & \underline{107.4} & 64 & \underline{92.2} \\
          \midrule[0.3pt]

          \multirow{2}{*}{SafeDrive}
          & $\times$
          & 27.5 & 41 & \underline{79.3} & 63.7 & 69.4 & 47
          & 30.4 & 45 & \underline{78.8} & 62.8 & 86.3 & 40 & 90.8 \\
          & $\checkmark$
          & \textbf{44.4} & 82 & 75.0 & \underline{69.8} & 68.3 & 44
          & \textbf{48.9} & 79 & 73.8 & \underline{69.5} & 81.3 & 49 & 90.6 \\
          \midrule[0.85pt]

          \multicolumn{15}{c}{\textit{VLA-based planners}} \\
          \midrule[0.3pt]

          \multirow{2}{*}{ReCogDrive}
          & $\times$
          & 15.4 & 71 & 75.7 & 59.0 & 76.9 & \textbf{100}
          & 20.3 & 62 & 77.1 & 58.7 & 93.7 & \underline{99} & 90.8 \\
          & $\checkmark$
          & 21.5 & 83 & 74.4 & 55.8 & 59.3 & 98
          & 26.3 & 84 & 75.6 & 56.7 & 75.5 & 98 & 90.4 \\

          \bottomrule[1.3pt]
      \end{tabular}%
      }
      \vspace{-5pt}
  \end{table*}

% \subsection{Simulator Validation}
% \textbf{\scshape Reconstruction Quality.} We assess the fidelity of Odyssey's rendered
% observations through the cumulative ablation in Table~\ref{tab:reconstruction_quality}.
% Starting from OmniRe~\citep{omnire}, the out-of-box loss and color correction adopted
% from MTGS~\citep{mtgs} improve reconstruction quality by constraining actor Gaussians
% and compensating for exposure variations. Diffusion refinement yields further gains,
% particularly after fine-tuning on Odyssey. Figure~\ref{fig:gs_comparison} shows the
% corresponding qualitative effect, where refined observations exhibit less blur and
% fewer rendering artifacts.

% \textbf{\scshape Traffic-Light Reconstruction.} Faithful simulation also requires capturing
% changes in traffic-light states, which the baseline reconstruction does not explicitly
% model. Odyssey addresses this limitation through a time-conditioned deformation field,
% reproducing the signal changes shown in Figure~\ref{fig:traffic} and enabling
% evaluation of planners' responses to the rendered signals.

\vspace{10pt}
\subsection{Effect of Diffusion Refinement on Closed-Loop Planning}
% Table~\ref{tab:refinement_ablation} evaluates visual refinement
%   with the reconstructed scene fixed, using the same SafeDrive
%   checkpoint with IL scoring and safety filtering.
%   Open-loop mIoU uses the logged trajectory, while closed-loop
%   mIoU compares all variants at identical camera poses from
%   the rollout with fine-tuned refinement.
%   Diffusion alone improves mIoU but slightly lowers \textit{RouteDS}.
% Odyssey-specific fine-tuning achieves the best results
%   in both metrics, improving scene interpretation and driving performance.

Table~\ref{tab:refinement_ablation} compares visual refinement
  settings with the reconstructed scene and SafeDrive checkpoint
  fixed, using IL scoring and safety filtering.
  Open-loop mIoU is evaluated along the logged trajectory.
  For closed-loop mIoU, all variants use identical camera poses
  from the rollout with fine-tuned refinement.
Pretrained Fixer improves mIoU, but these gains are
  accompanied by a slight decrease in RouteDS.
  Fine-tuning on Odyssey further improves mIoU in both
  settings, brings open-loop mIoU closer to that obtained
  with recorded images, and raises RouteDS above
  the GS-only baseline.
  These results support adapting the refinement model
  to Odyssey's reconstruction artifacts to improve both
  perception and closed-loop driving performance.

% \begin{figure}[t]
%     \centering
%     \begin{minipage}[t]{0.49\textwidth}
%       \vspace{0pt}
%       \centering
%       \includegraphics[width=\linewidth]{figure/Traffic_Light_v4.pdf}
%       \caption{\textbf{Traffic-light modeling.}
%       Our deformation field reproduces the red-to-green
%       transition, which OmniRe fails to recover.}
%       \label{fig:traffic}
%     \end{minipage}
%     \hfill
%     \begin{minipage}[t]{0.49\textwidth}
%       \vspace{0pt}
%       \centering
%       \includegraphics[width=\linewidth]{figure/Qualitative_Command_vs_SD.pdf}
%       \caption{\textbf{Driving commands vs. SD-route guidance.}
%       With driving commands, SafeDrive deviates from the intended
%       route when the command flips.
%       With SD-route guidance, it follows the intended route.}
%       \label{fig:command_sd}
%     \end{minipage}
%   \end{figure}

\subsection{Effect of SD-Route Guidance on Closed-Loop Driving}
% Table~\ref{tab:sdroute_baselines} compares planners with and without
%   SD-route guidance under non-reactive and reactive traffic.
%   SD-route guidance consistently improves SDC across planners,
%   indicating better adherence to the intended route.
%   This improvement is illustrated in Figure~\ref{fig:command_sd},
%   where SafeDrive deviates from the intended route when the driving
%   command flips but follows it with SD-route guidance.
%   The table also shows that higher NAVSIM open-loop PDMS does not
%   consistently correspond to higher RouteDS.
%   This mismatch highlights the need for continuous closed-loop
%   evaluation, where planning decisions affect subsequent observations
%   and outcomes.

Table~\ref{tab:sdroute_baselines} compares planners with and
  without SD-route guidance under non-reactive and reactive traffic.
  Providing SD routes improves both SDC and RouteDS across all
  evaluated planners in both 
  \begin{wrapfigure}{r}{0.49\textwidth}
  \vspace{0pt}
  \centering
  \includegraphics[width=\linewidth]{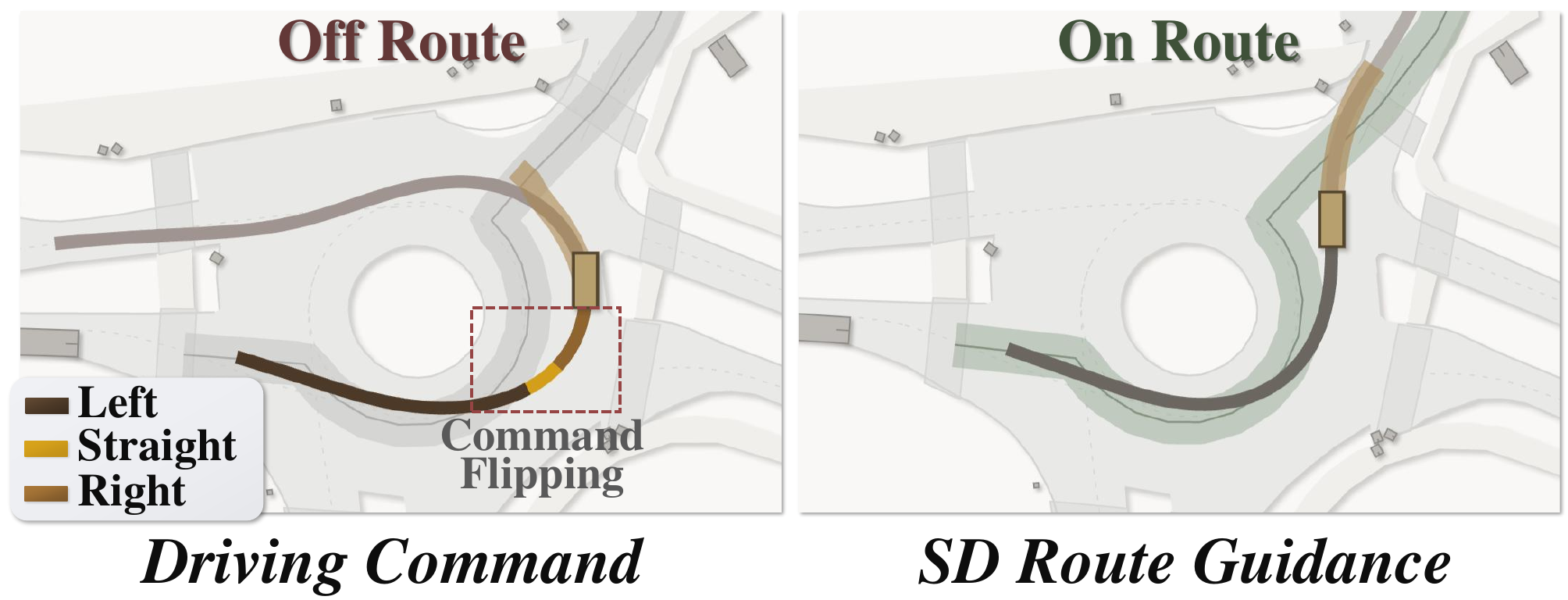}
  \caption{\textbf{Driving commands vs. SD-route guidance.}
  With driving commands, SafeDrive deviates from the intended
  route when the command flips.
  With SD-route guidance, it follows the intended route.}
  \label{fig:command_sd}
  \vspace{-10pt}
\end{wrapfigure} 
settings.
  Figure~\ref{fig:command_sd} illustrates this effect through
  SafeDrive, which deviates from the intended route when the
  driving command flips but follows it with SD-route guidance.

  Even with explicit route guidance, planners differ in their
  ability to combine route adherence with safe driving.
  DiffusionDrive achieves higher SDC than SafeDrive, but
  SafeDrive obtains higher RouteDS, which additionally accounts
  for route progress and driving infractions.
  NAVSIM PDMS also does not consistently reflect these
  differences in closed-loop performance.
  For example, DrivoR achieves higher PDMS than SafeDrive
  but lower RouteDS in both traffic settings.
  These comparisons motivate evaluating how planners follow
  the intended route over successive decisions, including
  the driving situations resulting from their own actions.

% \vspace{-5pt}

\begin{table}[!t]
  \centering
  \caption{Effects of safety scoring and reinforcement learning
  on driving performance under non-reactive and reactive traffic
  and on NAVSIM.
  All models use SD-map route guidance.}
  \label{tab:planner_variants}
  % \vspace{-5pt}
  \renewcommand{\arraystretch}{0.95}
  \setlength{\tabcolsep}{3pt}
  \resizebox{\textwidth}{!}{%
  \begin{tabular}{@{}l | c c c c c c | c c c c c c | c@{}}
      \toprule[1.3pt]

      \multirow[c]{2}{*}[-0.5ex]{\textbf{Configuration}}
      & \multicolumn{6}{c|}{\textbf{Odyssey Non-reactive}}
      & \multicolumn{6}{c|}{\textbf{Odyssey Reactive}}
      & \textbf{NAVSIM} \\

      \cmidrule(lr){2-7}
      \cmidrule(lr){8-13}
      \cmidrule(lr){14-14}

      & \textit{\textbf{RouteDS}}
      & \textbf{SDC}
      & \textbf{PLCA}
      & \textbf{PLCS}
      & \textbf{Eff.}
      & \textbf{Comf.}
      & \textit{\textbf{RouteDS}}
      & \textbf{SDC}
      & \textbf{PLCA}
      & \textbf{PLCS}
      & \textbf{Eff.}
      & \textbf{Comf.}
      & \textbf{PDMS} \\
      \midrule[0.85pt]

      \multicolumn{14}{c}{\textit{SafeDrive}} \\
      \midrule[0.3pt]

            Safety Scoring
      & 44.4 & 82 & 75.0 & 69.8 & \textbf{68.3} & 44
      & 48.9 & 79 & 73.8 & 69.5 & \textbf{81.3} & 49 & \textbf{90.6} \\

      IL Scoring
      & 53.9 & 95 & \textbf{77.1} & \textbf{74.4} & 66.5 & \textbf{100}
      & 55.3 & 94 & \textbf{78.0} & \textbf{74.4} & 75.7 & \textbf{100} & 89.5 \\

      IL Scoring + Safety Filtering
      & \textbf{54.4} & \textbf{96} & 76.7 & 72.7 & 65.3 & \textbf{100}
      & \textbf{56.1} & \textbf{96} & 77.7 & 74.1 & 78.3 & \textbf{100} & 89.5 \\
      \midrule[0.85pt]

      \multicolumn{14}{c}{\textit{ReCogDrive}} \\
      \midrule[0.3pt]

      IL + GRPO Fine-tuning
      & 21.5 & 83 & 74.4 & 55.8 & \textbf{59.3} & 98
      & 26.3 & 84 & 75.6 & 56.7 & \textbf{75.5} & 98 & \textbf{90.4} \\

      IL only
      & \textbf{37.7} & \textbf{97} & \textbf{79.4} & \textbf{73.8} & 58.9 & \textbf{100}
      & \textbf{42.3} & \textbf{96} & \textbf{79.6} & \textbf{75.0} & 70.4 & \textbf{100} & 86.8 \\

      \bottomrule[1.3pt]
  \end{tabular}%
  }
  \end{table}

% \begin{table}[t]
%   \centering

%   \begin{minipage}[t]{0.54\linewidth}
%   \vspace{0pt}
%   \centering
%   \caption{\textbf{Effect of safety scoring on SafeDrive.}
%   All variants share the same checkpoint and differ only
%   in trajectory selection.}
%   \label{tab:safety_scoring}
%   \renewcommand{\arraystretch}{0.95}
%   \setlength{\tabcolsep}{4pt}
%   \resizebox{\linewidth}{!}{%
%   \begin{tabular}{c | c c c | c c c | c}
%   \toprule[1.3pt]
%   \multirow[c]{2}{*}[-0.5ex]{\textbf{Trajectory Selection}}
%   & \multicolumn{3}{c|}{\textbf{Non-reactive}}
%   & \multicolumn{3}{c|}{\textbf{Reactive}}
%   & \textbf{NAVSIM} \\
%   \cmidrule(lr){2-4} \cmidrule(lr){5-7}
%   & \textit{\textbf{RouteDS}} & \textbf{RC} & \textbf{P$_{\mathrm{col}}$}
%   & \textit{\textbf{RouteDS}} & \textbf{RC} & \textbf{P$_{\mathrm{col}}$}
%   & \textbf{PDMS} \\
%   \midrule[0.85pt]
%   Safety scoring        & - & - & - & - & - & - & - \\
%   IL score only         & - & - & - & - & - & - & - \\
%   IL + Safety filtering & - & - & - & - & - & - & - \\
%   \bottomrule[1.3pt]
%   \end{tabular}%
%   }
%   \end{minipage}
%   \hfill
%   \begin{minipage}[t]{0.45\linewidth}
%   \vspace{0pt}
%   \centering
%   \includegraphics[width=\linewidth]{figure/Qualitative_Command_vs_SD.pdf}
%   \captionof{figure}{\textbf{SD route.}
%   The route specifies the intended road.}
%   \label{fig:sd_route}
%   \end{minipage}

%   \end{table}

\subsection{Do Open-Loop Gains Transfer to Closed-Loop Driving?}
Recent research has explored trajectory scoring and
  reinforcement learning as ways to improve driving safety.
  In NAVSIM-based training, trajectory scorers learn to predict
  simulation-based driving scores, while reinforcement learning
  uses these scores as rewards.
  Both objectives evaluate individual trajectories in an
  open-loop setting, without directly optimizing how the
  planner responds to new observations after execution.

  Table~\ref{tab:planner_variants} shows the limits of this approach.
  For SafeDrive, IL scoring outperforms safety scoring despite
  similar PDMS values.
  Safety filtering further improves performance by selecting
  the highest IL-scored trajectory among candidates that pass
  safety checks.
  For ReCogDrive, GRPO improves PDMS but lowers closed-loop
  performance relative to IL-only training.
  These results suggest that optimizing individual trajectories
  does not ensure better continuous driving.
  Odyssey closes this gap by evaluating planners on the consequences of their own decisions, not on isolated trajectories.

\subsection{Why Evaluate over Long Horizons?}
% Short-horizon evaluation tests whether a planner can handle
%   an immediate driving situation, but may not reveal whether
%   it can sustain good performance over successive maneuvers.
%   Figure~\ref{fig:analysis}(a) plots each model's RouteDS
%   against the evaluation horizon.
%   RouteDS decreases as the horizon grows for all planners,
%   reflecting the difficulty of sustaining performance over
%   longer rollouts.
%   More importantly, the curves cross: DrivoR leads at short
%   horizons but falls behind SafeDrive as driving continues.
%   Short-horizon results therefore do not fully capture a
%   planner's ability to sustain safe, route-following behavior,
%   which long-horizon evaluation reveals by testing successive
%   decisions and the situations they create.

\begin{figure}[!t]
\centering
\includegraphics[width=\columnwidth]{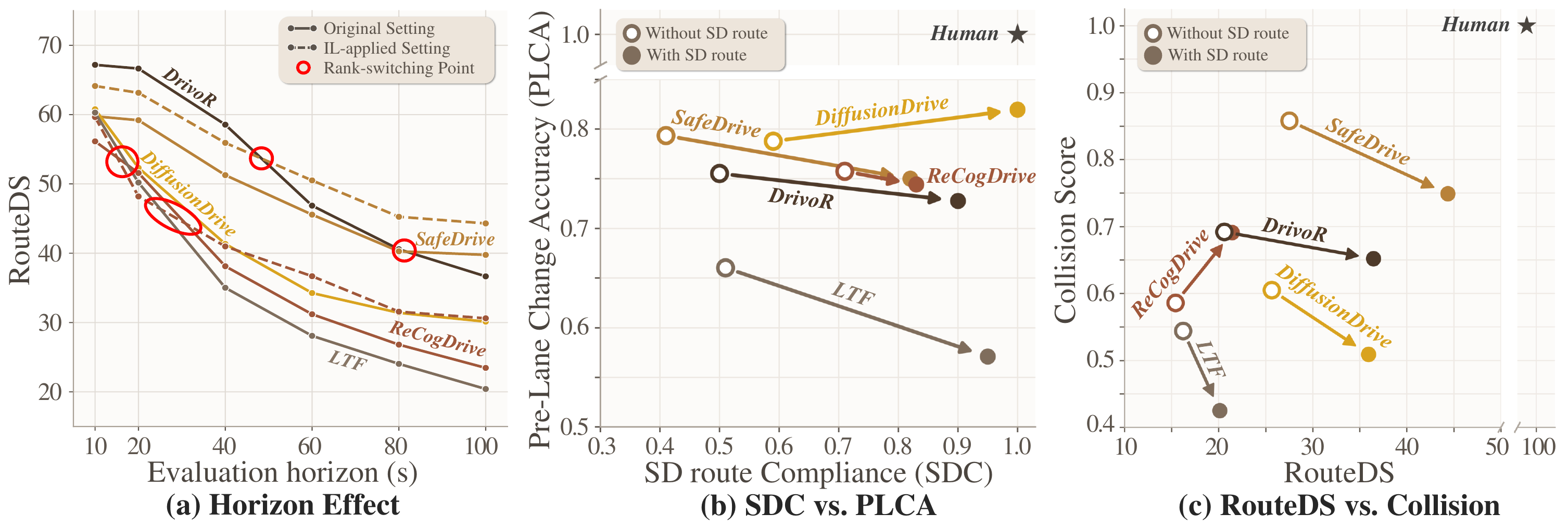}
% \vspace{-20pt}
\caption{\textbf{Effects of evaluation horizon and route guidance 
  under non-reactive traffic.}
  (a) RouteDS decreases with longer evaluation horizons,
  and model rankings change at the highlighted crossings.
  Solid and dashed curves show the original and IL-based
  settings, respectively.
  (b) SD-route guidance improves SDC but reduces PLCA
  for most planners.
  (c) RouteDS improves with SD-route guidance, but collision
  scores decrease for most planners.
  Higher collision scores indicate fewer collision penalties.
  }
% \vspace{-10pt}
\label{fig:analysis}
\end{figure}

Short-horizon evaluation assesses immediate driving behavior
  but may not reveal whether a planner can sustain performance
  over successive decisions.
  Figure~\ref{fig:analysis} (a) shows that RouteDS decreases
  as the evaluation horizon grows, highlighting the challenge
  of sustaining safe, route-following behavior over longer rollouts.
  Some decline is expected as infraction penalties accumulate,
  but the relative performance of planners also changes.
  DrivoR achieves higher RouteDS than SafeDrive at short
  horizons but falls below it as driving continues.
  This rank reversal shows that conclusions drawn from short
  rollouts may not hold over longer ones, motivating evaluation
  of successive decisions and their downstream consequences.

\subsection{What Is Needed Beyond Route Guidance?}

% Figure~\ref{fig:analysis}(b) shows that replacing driving commands
%   with SD-route guidance substantially improves SDC, whereas PLCA
%   improves only slightly, remains similar, or decreases.
%   Knowing the intended roads thus helps models follow the route,
%   but not position themselves in the lanes needed for upcoming
%   maneuvers.
%   Figure~\ref{fig:analysis} (c) reveals a similar gap in safety.
%   Although SD-route guidance raises RouteDS for every planner,
%   the collision score decreases for four of the five.
%   Route guidance alone therefore ensures neither lane-level
%   preparation nor safe interaction with surrounding traffic,
%   and closing both gaps remains an open direction for future
%   research on Odyssey.

Figure~\ref{fig:analysis} (b) shows that SD-route guidance
  improves SDC for all planners, but PLCA decreases for
  four of the five, with only DiffusionDrive showing an increase.
  The gains in road-level route adherence therefore do not
  extend to timely lane positioning for most evaluated planners.
  Translating the intended road sequence into appropriate
  lane choices before intersections remains a challenge.

  Beyond lane preparation, following a route also requires
  safe interaction with surrounding traffic.
  Figure~\ref{fig:analysis} (c) shows that SD-route guidance
  raises RouteDS for every planner, but the collision score
  decreases for four of the five.
  The higher overall scores therefore do not consistently
  coincide with fewer collision penalties.
  These results highlight remaining challenges in lane
  preparation and traffic interaction under explicit route guidance.
  How planners can combine SD-map routes with perceived road
  geometry and surrounding traffic to address these challenges
  remains an open direction for future research on Odyssey.

\section{Future Works}
Surrounding vehicles can appear blurred or distorted when
  viewed from angles not covered by the driving logs.
  Using diffusion guidance during 3D Gaussian Splatting
  optimization could improve reconstruction in these regions.
  Future work could also improve the temporal consistency
  of diffusion refinement. Beyond visual quality, adversarial agents could introduce
  more challenging traffic interactions for safety evaluation.

% \begin{figure}[t]
%   \centering

%   \begin{minipage}[t]{0.40\linewidth}
%   \vspace{0pt}
%   \centering
%   \includegraphics[width=\linewidth]{figure/Traffic_Light_v4.pdf}
%   \caption{\textbf{Traffic-light modeling.}
%   Our deformation field reproduces the red-to-green transition,
%   which OmniRe fails to recover.}
%   \label{fig:traffic}
%   \end{minipage}
%   \hfill
%   \begin{minipage}[t]{0.59\linewidth}
%   \vspace{0pt}
%   \centering
%   \includegraphics[width=\linewidth]{figure/Qualitative_Command_vs_SD.pdf}
%   \caption{\textbf{SD route.}
%   The route specifies the intended road.}
%   \label{fig:sd_route}
%   \end{minipage}
% \end{figure}

% \begin{wrapfigure}{r}{0.6\textwidth}
%     \vspace{-10pt}
%     \centering
%     \includegraphics[width=\linewidth]{figure/Qualitative_Command_vs_SD.pdf}
%     \vspace{-18pt}
%     % \caption{\textbf{Limitations of command-based guidance.} (a) The same command can indicate different maneuvers, leaving navigation intent ambiguous. (b) Extracted commands can flip inconsistently along a continuous trajectory.}
%     % \caption{\textbf{Limitations of existing evaluation setups.} (a) Short rollouts cannot assess whether an early lane change enables a later right turn. (b) Inconsistent command flips obscure navigation intent along a continuous trajectory.}
%     \caption{\textbf{Command.} }
%     \label{fig:command}
%     \vspace{-10pt}
% \end{wrapfigure}

\vspace{-5pt}
\section{Conclusion}
\vspace{-5pt}
% We introduced Odyssey, a closed-loop benchmark for long-horizon
%   driving in environments reconstructed from real-world logs.
%   By pairing continuous 100-second scenarios with explicit SD-map
%   routes, Odyssey evaluates how well planners sustain safe,
%   route-following driving through RouteDS, SDC, and PLCA.
%   Using this benchmark, we find that planner rankings shift as the
%   horizon grows and that open-loop gains do not necessarily carry
%   over to closed-loop driving.
%   We further find that SD-route guidance improves route adherence
%   but does not ensure timely lane positioning or safe interaction
%   with traffic.
%   These findings suggest that driving ability emerges from successive
%   decisions, and we hope Odyssey serves as a testbed for planners
%   that can sustain it.

  We introduced Odyssey to evaluate whether E2E planners can
  sustain safe navigation under explicit route guidance.
  Odyssey provides SD-map routes and continuous sensor feedback
  in environments reconstructed from 100-second real-world
  driving logs.
  This setting supports closed-loop evaluation under a
  consistent navigation objective.
  Our experiments show that planner rankings change with
  the evaluation horizon and that improvements in open-loop
  metrics do not necessarily translate into better closed-loop
  driving.
Explicit route guidance improves road-level adherence,
  but these gains do not consistently extend to lane preparation
  or collision avoidance.
  Odyssey supports research on these remaining challenges
  by evaluating how planners use route guidance and sensor
  observations to select appropriate lanes and interact safely
  with surrounding traffic over extended closed-loop rollouts.

% We introduced Odyssey, a closed-loop benchmark for long-horizon driving in reconstructed real-world environments. Odyssey pairs continuous 100-second scenarios with explicit SD-map routes to evaluate driving safety and sustained route progress. Diffusion-refined 3D Gaussian rendering provides visual feedback, while RouteDS penalizes departures from the intended route. Together with adapted E2E and VLA baselines, Odyssey provides a framework for examining the consequences of successive driving decisions.

\newpage
\clearpage

\bibliography{iclr2027_conference}
\bibliographystyle{iclr2027_conference}

\clearpage
\appendix
\startcontents[appendix]

% \suppressfloats[t]

  \begin{center}
      {\LARGE\scshape Appendix for Odyssey}
  \end{center}
  \vspace{0.5em}
  
  \noindent{\large\bfseries Contents}
  \par\vspace{0.5em}

\printcontents[appendix]{}{1}{%
    \setcounter{tocdepth}{2}%
}

  % \clearpage

  \section{Related Work}

\subsection{End-to-End Driving Models}
Planning-oriented E2E driving integrates perception, prediction, and planning~\citep{stp3, uniad},
  with subsequent work improving scene representation and task interaction~\citep{vad, sparsedrive,
  paradrive, drivetransformer}. Building on these representations, planners jointly predict
  trajectories and controls~\citep{tcp, thinktwice} or separate path geometry from speed
  prediction~\citep{transfuserpp, hipad}. To handle multiple feasible actions, diffusion models
  generate diverse trajectories~\citep{diffusiondrive}, and candidate-based planners learn to score
  alternatives~\citep{hydramdp, drivesuprim}. World models further inform planning through predicted
  future states and action risks~\citep{wote, safedrive}, and provide imagined interactions for
  reinforcement learning~\citep{think2drive}.

  Vision-language-action (VLA) models extend sensor-based planning with language understanding and
  reasoning. SimLingo aligns language instructions with driving actions~\citep{simlingo}, and
  DriveVLA-W0 incorporates future-image prediction to enrich visual learning beyond action
  supervision~\citep{drivevlaw0}. ReCogDrive and AutoDrive-R$^2$ further combine reasoning with
  reinforcement learning to improve driving decisions~\citep{recogdrive, autodriver2}. Odyssey adapts
  vision-based and VLA planners to SD-route guidance for evaluating sustained
  navigation.

\subsection{Evaluation of E2E Planning}

  Evaluating E2E planners requires assessing the consequences
  of their actions, which open-loop displacement-based metrics
  overlook~\citep{nuscenes, stp3, uniad}.
  NAVSIM~\citep{navsim} moves beyond imitation accuracy through
  simulation-based driving metrics.
  NAVSIM v2~\citep{navsimv2} further incorporates pre-rendered
  observations to approximate subsequent feedback.
  Extending this feedback to continuous replanning requires
  observations that reflect the ego vehicle's evolving state.
  Graphics-based simulators provide such observations but
  introduce a visual domain gap~\citep{carla, metadrive, bench2drive}.
  Planning simulators using real logs evaluate driving behavior
  without sensor rendering~\citep{nuplan1, nuplan2, waymax}.

Neural-rendering-based simulators bring sensor feedback to recorded driving environments by
  reconstructing scenes and rendering observations in response to ego actions~\citep{unisim, neurad,
  alpasim}. Within this paradigm, NeuroNCAP~\citep{neuroncap} examines collision avoidance in NeRF-reconstructed scenarios, and HUGSIM~\citep{hugsim} provides closed-loop evaluation through 3DGS
  rendering and controllable traffic. However, deviations from the logged trajectory can expose
  poorly reconstructed regions~\citep{drivinggaussian, hugsim, splatad, omnire}, producing artifacts that compromise the reliability of sensor
  feedback. Odyssey addresses this issue by refining 3DGS-rendered observations with diffusion to
  reduce artifacts at novel ego poses.

  \subsection{Driving Intention for End-to-End Models}
  Navigation guidance specifies the intended path when multiple
  driving maneuvers are feasible~\citep{cil}. Benchmarks based on
  real driving logs~\citep{uniad, vad, navsim, navsimv2, neuroncap, hugsim,
  wod_e2e} commonly express this intent through driving
  commands, but this compact representation cannot uniquely
  distinguish consecutive branches sharing the same direction.
  CARLA-based benchmarks~\citep{carla}, including Bench2Drive~\citep{bench2drive}, provide explicit
  guidance through target points or route waypoints in synthetic
  environments.
    However, guidance limited to nearby targets provides insufficient
  route context for planning across multiple maneuvers over
  longer horizons.
      Odyssey uses OSM-derived SD-map routes to provide this guidance
      in reconstructed real-world environments. These routes specify
      the intended road sequence throughout the rollout and reduce
      reliance on detailed HD maps as planner inputs.
      Local driving decisions remain based on sensor observations.

\section{Implementation Details of Closed-loop Simulator}

% In this section, we present the implementation details of the rendering component in our closed-loop simulator. We first describe the 3D Gaussian reconstruction framework, including Gaussian parameterization, data preprocessing, traffic-light modeling, and training objectives. We then detail the fine-tuning procedure for Fixer~\citep{fixer}, focusing on training-pair generation and optimization settings. Finally, we introduce the evaluation protocol and metrics used to assess rendering quality, perceptual fidelity, and geometric accuracy.

\subsection{3D Gaussian Reconstruction}

\paragraph{\scshape Reconstruction Framework.}
We build on OmniRe~\citep{omnire} to jointly reconstruct
  the static background and dynamic actors, and model traffic
  lights separately to reproduce signal changes.
  Pose noise in nuPlan logs degrades 3D Gaussian reconstruction
  quality, motivating pose refinement with COLMAP~\citep{colmap}.
  We keep the logged $x$ and $y$ coordinates fixed to preserve
  alignment with the HD map used for drivable-area evaluation,
  and optimize only $z$, roll, pitch, and yaw.
  To separate scene components, we generate sky and dynamic-object
  masks using SegFormer~\citep{segformer}.
  We initialize background Gaussians by back-projecting
  MoGe-2 pseudo-depth~\citep{moge2} after excluding dynamic pixels.
  Rigid and deformable actors are initialized from accumulated
  LiDAR points, and pedestrians from SMPL templates~\citep{loper2015smpl}.
  For traffic-light modeling, we obtain head masks using
  Mask2Former~\citep{mask2former} pretrained on Mapillary Vistas v2~\citep{mapillary}.

\paragraph{\scshape Ground Modeling.}
Inspired by HUGSIM~\citep{hugsim}, we represent the road surface
with a separate set of thin Gaussian disks.
We first accumulate LiDAR points using the refined ego poses and
build a road height field on a 0.125\,m grid.
Ground Gaussians are placed on this height field within the HD-map
road regions; heights of unobserved road cells are filled by
harmonic interpolation from the observed cells.
Each disk has a fixed thickness of 0.01\,m and stays attached to
the surface: its height and normal are periodically re-sampled from
the height field at its current horizontal position.
The remaining attributes are optimized, namely the horizontal position,
the in-plane scale, the rotation about the surface normal,
the opacity, and the spherical-harmonic coefficients.
To keep the background from also representing the road, we remove
background points within 0.3\,m of the surface at initialization.
During training, we also penalize the mean opacity of background
Gaussians that lie between 0.3\,m below and 1.0\,m above the surface.

\paragraph{\scshape Coverage-aware Sampling.}
Sampling images uniformly over a long driving sequence concentrates
training updates on places where the recording vehicle stops or moves
slowly.
To balance spatial coverage, we back-project the LiDAR depth of each
training image, excluding dynamic objects, into 0.5\,m voxels and count
how many training views observe each voxel.
The weight of an image is the mean inverse observation count of its
visible voxels, normalized by the 90th percentile over all images and
clipped to $[0.1, 1]$.
Images that mainly see less-observed regions therefore receive higher
weights, and frequently observed regions are revisited less often.
Rather than drawing images at random, we build a sampling cycle in
which each image appears a number of times proportional to its weight,
so that every training view is used at least once per cycle.
Images containing dynamic objects are given a higher minimum weight of
0.25 to preserve the temporal coverage of moving actors.
Each reconstruction update uses a single image.

\paragraph{\scshape Traffic-Light Modeling.}
To capture signal changes, we represent each traffic-light head as a
separate deformable Gaussian node.
Using head masks predicted by Mask2Former~\citep{mask2former} and depth
estimated by MoGe-2~\citep{moge2}, we back-project masked pixels from
calibrated views to initialize Gaussian positions and colors.
Each node has a fixed installation position and orientation, and its
geometry is kept fixed up to a small tolerance of 5\,cm that absorbs
residual calibration error.
Signal changes are therefore modeled through appearance: an MLP takes
Gaussian positions, time, and a learned embedding for each traffic
light, and predicts temporal residuals to the spherical-harmonic
coefficients and opacity.
After reconstruction, we manually verify the signal schedules against the rendered traffic-light states and revise them where necessary.

 \begin{table}[!t]
      \centering
      \small
      \setlength{\tabcolsep}{5pt}
      \renewcommand{\arraystretch}{0.95}

      \caption{Gaussian densification, pruning, and optimization settings.}
      \label{tab:gaussian_optimization}

      \begin{tabular}{@{}lcc@{}}

        \toprule
        \textbf{Parameter}
        & \textbf{Background}
        & \textbf{Rigid actors} \\
        \midrule

        \multicolumn{3}{c}{\textit{Densification}} \\
        \specialrule{0.3pt}{0.0ex}{0.15ex}

        Start / end iteration
        & $2{,}500$ / $93{,}500$
        & $2{,}500$ / $93{,}500$ \\
        Interval (iterations)
        & $500$ & $500$ \\
        Views per scoring step
        & $120$ & $120$ \\
        Normalized error threshold
        & $0.1$ & $0.1$ \\
        Required score
        & $>5$ & $>5$ \\
        Clone gradient threshold
        & $2\times10^{-4}$ & $2\times10^{-4}$ \\
        Split absolute-gradient threshold
        & $1\times10^{-4}$ & $3\times10^{-4}$ \\
        Clone / split scale boundary
        & $0.0562\,\mathrm{m}$ & $0.09\,\mathrm{m}$ \\
        Children per split
        & $2$ & $2$ \\
        \midrule

        \multicolumn{3}{c}{\textit{Pruning and opacity control}} \\
        \specialrule{0.3pt}{0.0ex}{0.15ex}

        Minimum views for pruning
        & $5$ & $2$ \\
        Early opacity threshold
        & $0.005$ & $0.005$ \\
        Early scale threshold
        & $5.617\,\mathrm{m}$ & $5.617\,\mathrm{m}$ \\
        Opacity cap
        & $0.8$ & $0.8$ \\
        Opacity reset interval
        & $7{,}000$ & $14{,}500$ \\
        Opacity reset value
        & $0.01$ & $0.01$ \\
        Post-pruning start / end
        & $70{,}000$ / $140{,}000$
        & $93{,}500$ / $140{,}000$ \\
        Post-pruning interval
        & $4{,}500$ & $4{,}500$ \\
        Post-pruning opacity threshold
        & $0.05$ & $0.05$ \\
        Post-pruning score threshold
        & $0.9$ & $0.9$ \\
        \midrule

        \multicolumn{3}{c}{\textit{Background update intervals (iterations)}} \\
        \specialrule{0.3pt}{0.0ex}{0.15ex}

        \textbf{Training iterations}
        & \textbf{Geometry / opacity} & \textbf{SH} \\
        Up to $70{,}000$ & $1$ & $16$ \\
        $70{,}001$--$93{,}334$ & $32$ & $32$ \\
        After $93{,}334$ & $64$ & $64$ \\
        \bottomrule
      \end{tabular}
  \end{table}

  \paragraph{\scshape Optimization.}
  % We jointly optimize each complete 100-second sequence
  % as a single scene for approximately $140\,\mathrm{k}$ iterations,
  % sampling one training image per iteration.
  % We use eight cameras at $10\,\mathrm{Hz}$ with
  % $1920\times1080$ images, starting at half resolution
  % and switching to full resolution at iteration $1{,}166$.
  % To balance training across the sequence, we assign higher
  % sampling weights to views covering less frequently observed
  % $0.5\,\mathrm{m}$ spatial voxels, retaining every training
  % view in each sampling cycle.
  % FastGS~\citep{fastgs}, GPU-resident image and supervision
  % caching, and less frequent background parameter updates
  % reduce training overhead.
  We jointly optimize each complete 100-second sequence as a single scene for 140k iterations, sampling one training image per iteration. We use eight cameras at 10 Hz with 1920×1080 images, starting at half resolution and switching to full resolution at iteration 1,166. To balance training across the sequence, we assign higher sampling weights to views covering less frequently observed 0.5 m spatial voxels, while every training view is still served at least once per sampling cycle. FastGS, GPU-resident caching of images and supervision signals, and a reduced update cadence for the background parameters of Gaussians reduce training overhead.

We apply FastGS density control~\citep{fastgs} to background
  and rigid-actor Gaussians, computing scores from 120 views
  sampled across cameras and the full sequence.
  To account for uneven visibility over long sequences,
  we normalize densification scores by each Gaussian's
  visible-view count.
  Table~\ref{tab:gaussian_optimization} lists the settings.

\paragraph{\scshape Training Objectives.}
We jointly optimize appearance and geometry using a weighted
  sum of the losses.
  Our appearance objective combines RGB L1 and SSIM losses
  with color-correction regularization.
  RGB L1 compares color-corrected renderings with recorded
  images, and SSIM supervises the uncorrected renderings
  to preserve image structure~\citep{vastgaussian}.
  The regularization penalizes colors outside the valid range,
  limits frame-dependent corrections, and discourages abrupt
  changes between adjacent frames.
  To account for exposure and white-balance differences,
  we adjust rendered RGB values using a learnable
  $3\times4$ affine transformation following MTGS~\citep{mtgs}.
The first three columns transform the RGB channels,
  and the last column adds a per-channel offset.
  A shared matrix captures persistent differences between
  cameras, and a residual matrix for each camera and frame
  captures temporal variations.

  Our geometry losses regularize scene structure and reduce
  reconstruction artifacts.
  LiDAR depth provides additional geometric supervision,
  and edge-aware smoothness reduces depth noise without
  smoothing across image boundaries~\citep{pvg}.
  Sky-opacity BCE encourages transparent sky regions,
  and a line-of-sight loss suppresses opacity at least
  $0.8\,\mathrm{m}$ in front of LiDAR returns~\citep{splatad}.
  Opacity entropy further encourages values near zero or one.
  For actors, we apply bounding-box regularization to both
  rigid and deformable Gaussians, extending OmniRe's
  regularization to rigid actors.
  We separately penalize local displacement of traffic-light
  Gaussians to limit deformation around their fixed installations. Table~\ref{tab:reconstruction_losses} summarizes the loss terms and their weights.

\begin{table}[!t]
      \centering
      \setlength{\tabcolsep}{6pt}
      \renewcommand{\arraystretch}{0.95}
      \caption{Reconstruction losses and weights.}
      \label{tab:reconstruction_losses}

      \begin{tabular}{@{}llc@{}}
          \toprule
          \textbf{Loss} & \textbf{Objective}
          & \textbf{Weight} ($\lambda$) \\
          \midrule[0.3pt]

          \multicolumn{3}{c}{\textit{Appearance supervision}} \\
          \specialrule{0.3pt}{0.0ex}{0.15ex}

          $\mathcal{L}_{\mathrm{rgb}}$
          & RGB L1 after color correction & 0.8 \\
          $\mathcal{L}_{\mathrm{ssim}}$
          & SSIM before color correction & 0.2 \\
          $\mathcal{L}_{\mathrm{color}}$
          & Out-of-range color penalty & 0.8 \\
          $\mathcal{L}_{\mathrm{res}}$
          & Frame-dependent residual magnitude & 0.01 \\
          $\mathcal{L}_{\mathrm{temp}}$
          & Temporal variation of color residuals & 0.1 \\
          \midrule

          \multicolumn{3}{c}{\textit{Geometry and opacity supervision}} \\
          \specialrule{0.3pt}{0.0ex}{0.15ex}

          $\mathcal{L}_{\mathrm{sky}}$
          & Sky-opacity BCE & 0.05 \\
          $\mathcal{L}_{\mathrm{depth}}$
          & LiDAR inverse-depth supervision & 0.1 \\
          $\mathcal{L}_{\mathrm{ent}}$
          & Opacity entropy & 0.05 \\
          $\mathcal{L}_{\mathrm{smooth}}$
          & Edge-aware inverse-depth smoothness & 0.001 \\
          $\mathcal{L}_{\mathrm{los}}$
          & Line-of-sight opacity & 0.1 \\
          $\mathcal{L}_{\mathrm{box}}$
          & Actor bounding-box regularization & 1.0 \\
          $\mathcal{L}_{\mathrm{TL}}$
          & Traffic-light Gaussian displacement & 1.0 \\

          \bottomrule
      \end{tabular}
  \end{table}

\subsection{Fixer fine-tuning}
We refine rendered images using Fixer~\citep{fixer},
  a single-step diffusion model built on the 0.6B
  Cosmos-Predict2 backbone~\citep{cosmos_predict2}.
  The autoencoder encodes the degraded rendering directly
  into latent features without adding random noise.
  The degraded rendering therefore serves as the input
  to be restored, with reconstruction artifacts providing
  the corruption that the model learns to remove.
  A diffusion transformer refines the latent features
  in one step, and the decoder reconstructs the output image.
  To adapt the pretrained model to Odyssey, we construct
  rendering--real image pairs from all 100 benchmark scenes
  using three degradation strategies following
  Difix3D+~\citep{difix3d}.

  \begin{itemize}
      \setlength{\itemsep}{4pt}
      \setlength{\parskip}{0pt}

      \item \textbf{\scshape Under-trained reconstruction (30\%).}
      We use renderings from the $80\,\mathrm{k}$,
      $120\,\mathrm{k}$, and final $140\,\mathrm{k}$
      checkpoints to cover different reconstruction quality levels.
      Final-checkpoint renderings discourage unnecessary
      corrections to high-quality inputs.

      \item \textbf{\scshape Cycle reconstruction (40\%).} %frame 마다 1m~6m shift 주입
      We render a trained Gaussian scene from a trajectory
      shifted laterally by $2\,\mathrm{m}$, train a second
      Gaussian model on these renderings, and render the
      original trajectory from the second model.
      These images are paired with the original real images
      to train refinement of novel-view artifacts.

      \item \textbf{\scshape Cross-reference reconstruction (30\%).}
      We train a Gaussian model using six of the eight cameras,
      holding out \texttt{CAM\_L2} and \texttt{CAM\_R2}.
      Renderings from the two held-out cameras are paired
      with their real images.
      These viewpoints receive only indirect supervision
      from neighboring cameras, producing novel-view artifacts.
  \end{itemize}

We randomly sample $20\%$ of the frames from the
  $10\,\mathrm{Hz}$ sequences for training-pair generation.
  Under-trained, cycle, and cross-reference reconstruction
  account for $30\%$, $40\%$, and $30\%$ of the training pairs,
  respectively.
  Pairs are generated at $576\times1024$ resolution
  without image-space augmentation.
  
  \paragraph{\scshape Training Objectives.}
  Using these pairs, we train Fixer with pixel, perceptual,
  and texture supervision:
  \begin{equation}
      \mathcal{L}_{\mathrm{Fixer}}
      =
      \lambda_{\mathrm{L2}}\mathcal{L}_{\mathrm{L2}}
      + \lambda_{\mathrm{LPIPS}}\mathcal{L}_{\mathrm{LPIPS}}
      + \lambda_{\mathrm{Gram}}\mathcal{L}_{\mathrm{Gram}}.
      \label{eq:fixer_loss}
  \end{equation}
    Full-image L2 loss measures pixel error,
  LPIPS~\citep{lpips} preserves perceptual content,
  and Gram loss~\citep{gatys} matches texture statistics.
  LPIPS and Gram losses use VGG-16 features from spatially
  aligned crops of the refined and real images.
  LPIPS uses square crops with side lengths of
  $128$--$512$ pixels, and Gram loss uses $512\times512$ crops.

  We set $\lambda_{\mathrm{L2}}=\lambda_{\mathrm{LPIPS}}=1$
  throughout training.
  Fixer's public training recipe disables Gram loss~\citep{fixer},
  but we found that activating it during fine-tuning substantially
  reduced blur and improved texture detail.
  To avoid excessive sharpening, we keep
  $\lambda_{\mathrm{Gram}}=0$ until the final $1{,}000$ steps,
  then set it to $0.02$.
  \paragraph{\scshape Inference Optimization.}
    To improve inference speed, we remove cross-attention layers inactive under zero text conditioning and employ TensorRT to accelerate Fixer's encoder, diffusion transformer, and decoder. These optimizations achieve a $3.7\times$ speedup with negligible changes in restoration quality  (Table~\ref{tab:fixer_runtime}).
    
\begin{table}[!t]
    \vspace{-4pt}
    \centering
    \setlength{\belowcaptionskip}{2pt}
    \caption{Fixer inference acceleration.}
    \label{tab:fixer_runtime}
    \small
    \setlength{\tabcolsep}{5pt}
    \begin{tabular}{@{}lcc@{}}
        \toprule
        Method & Time (ms/image) $\downarrow$ & Speedup $\uparrow$ \\
        \midrule
        Baseline  & 25.0 & $1.0\times$ \\
        Optimized & \textbf{6.7} & $\mathbf{3.7\times}$ \\
        \bottomrule
    \end{tabular}
    \par\vspace{-6pt}
\end{table}

\subsection{Section-based Traffic Triggering}

\begin{figure}[!t]
  \centering
  \includegraphics[width=\textwidth]{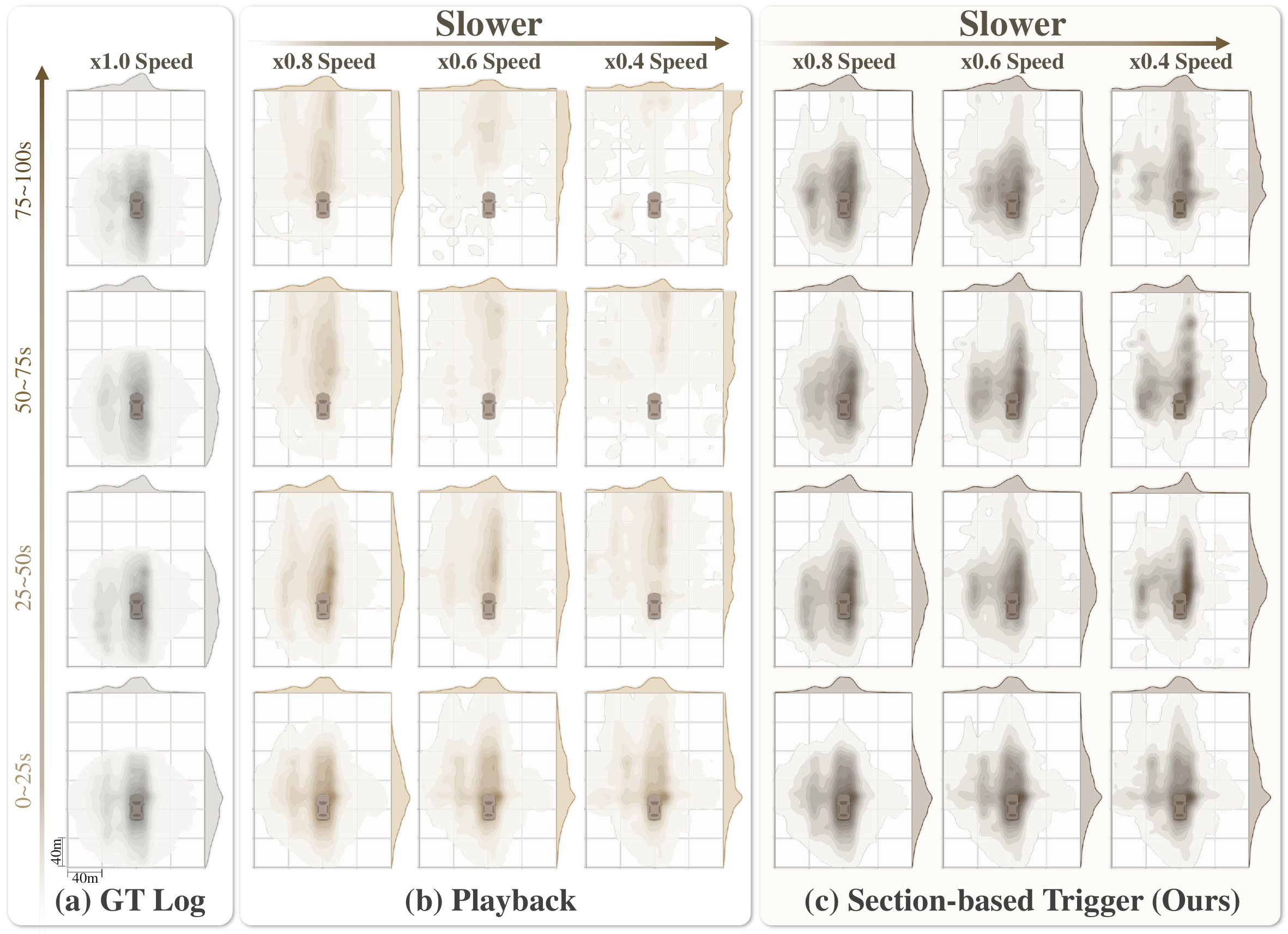}
\caption{\textbf{Effect of section-based triggering on nearby agent density.}
  (a) GT Log shows nearby agent density during recorded driving.
  (b) Playback follows the original traffic timeline, with density
  decreasing as the ego slows and simulation continues.
  (c) Section-based triggering aligns traffic activation with ego
  progress, maintaining density closer to GT Log across the
  evaluated speeds and time intervals.}
  \label{fig:trigger_appendix}
\end{figure}

When the simulated ego travels faster or slower than the logged
     vehicle, section-based traffic triggering aligns traffic playback
     with ego progress to preserve local traffic encounters.
     We place section boundaries between intersections and combine
     short sections to maintain a minimum length of 50\,m.

At each simulation step, we project the ego position onto the
  logged trajectory to determine its progress.
  The first section is active at initialization, and subsequent
  sections activate when the ego reaches 15\,m before their start.
  This 15\,m margin reduces sudden vehicle appearances at section
  boundaries and gives the planner time to observe the surrounding
  traffic before entering.
  To align each section's traffic with ego progress, we compute
  a fixed offset between the logged and simulated ego arrival
  times at the activation point.
  This offset provides a common timing reference for initializing
  surrounding vehicles and pedestrians and assigning traffic-light states.
  Parked vehicles and static objects remain active throughout
  the simulation, regardless of section activation.

  The two traffic settings differ in how surrounding vehicles
  move after initialization.
  In the non-reactive setting, vehicles replay their recorded
  motion using the section's fixed offset.
  In the reactive setting, vehicles are initialized with their
  logged positions and states, then follow IDM-based control.
  Pedestrian motion and traffic-light changes follow their
  recorded timing with the same section offset in both settings.

Figure~\ref{fig:trigger_appendix} compares nearby agent density
  under GT Log, Playback, and section-based triggering.
  With Playback, nearby agent density progressively decreases
  when the ego travels more slowly than the logged vehicle.
  Section-based triggering maintains density closer to GT Log
  across the evaluated ego speeds and time intervals.

  In addition to maintaining traffic density, reliable evaluation
  requires adequate visual quality for the surrounding actors.
  Distant or occluded actors may have too few observations
  for Gaussian reconstruction, resulting in poor rendering quality
  and insufficient visual cues for the planner.
  Such actors can still incur evaluation penalties.
  We therefore manually inspected individual actors and removed
  those with inadequate rendering quality.

% We provide implementation details for the triggering scheme
% introduced in Section~\ref{sec:traffic_simulation}.
% At each simulation step, we project the ego position onto
% the logged trajectory and retain the maximum arc-length
% progress reached so far.
% The first section is active at initialization; subsequent
% sections activate when this progress reaches 15 m before
% their start. Arrival at this activation point defines the
% entry time used to compute the section offset.
% In the reactive setting, vehicles are initialized to their
% first logged states and retained while within 50 m of the ego.
% Parked vehicles and static objects are excluded from triggering.

% We assign traffic signals to sections by projecting their
% installation positions onto the logged ego trajectory,
% including signals for crossing and opposing traffic.
% Planner inputs, IDM control, signal rendering, and red-light
% violation evaluation use the same section offset.
% Before section activation or outside the recorded time range,
% signal states are treated as unavailable and excluded from
% planner and IDM signal inputs and red-light violation
% evaluation. We do not extend the final recorded signal state,
% avoiding indefinite stops caused by a final red light.

  \section{Evaluation Metric Details}
  \label{app:metrics}

This section defines the rollout termination criteria,
  pre-lane change metrics, RouteDS components,
  and additional driving metrics.

  \subsection{Rollout Termination Criteria}
Each rollout evaluates progress toward a designated destination.
  Termination criteria bound its duration and prevent continued
  simulation far from the intended route. A rollout ends under
  one of three conditions:

  \begin{itemize}
      \setlength{\itemsep}{3pt}
      \setlength{\parskip}{0pt}

      \item \textbf{\scshape Destination arrival.}
      The ego reaches the end of the designated SD route.
      
      % The ego reaches the intended destination after progressing through the designated SD route.
      
      \item \textbf{\scshape Time limit.}
    The rollout is limited to $200\,\mathrm{s}$, twice the
      duration of the recorded ground-truth driving sequence.

      \item \textbf{\scshape SD-route deviation.}
    The rollout terminates as soon as the ego vehicle is more
  than $30\,\mathrm{m}$ from the designated SD route.
    
  \end{itemize}
  
\subsection{Pre-Lane Change Metrics}
We select evaluation intersections where the logged vehicle
  travels in route-compatible lanes throughout the
  $10\,\mathrm{m}$ before the stop line.
  To detect when vehicles approaching through oncoming lanes
  or off-road areas pass an evaluation intersection, we extend
  each stop line by $25\,\mathrm{m}$ at both ends and check
  for crossings.
  Crossings along these extensions, outside the original
  stop-line segment, are treated as failures to cross
  the stop line correctly.

  \subsection{RouteDS Components}
  For each scenario, we compute
  \begin{equation}
      \mathit{RouteDS}
      =
      100 \cdot \mathit{RC}_{\mathit{SD}}
      \cdot P_{\mathrm{PLC}}
      \cdot P_{\mathrm{SD}}
      \cdot P_{\mathrm{col}}
      \cdot P_{\mathrm{off}}
      \cdot P_{\mathrm{TL}},
      \label{eq:route_ds}
  \end{equation}
  where $\mathit{RC}_{\mathit{SD}}\in[0,1]$ denotes route
  completion and the remaining factors are penalties
  in $[0,1]$, defined below.
  We report the mean \textit{RouteDS} across scenarios.

  \begin{itemize}
      \setlength{\itemsep}{4pt}
      \setlength{\parskip}{0pt}

      \item \textbf{\scshape Route completion.}
       $\mathit{RC}_{\mathit{SD}}$ is the fraction of the designated
      SD route completed by the ego vehicle.
      After each rollout, we apply HMM map matching~\citep{newson2009hidden} to the executed trajectory to identify the SD-route segments traversed by the ego.
       If the ego deviates from the route, we determine where the ego
      left the route by repeating HMM matching from the start,
      extending the evaluated trajectory by $1\,\mathrm{m}$ each time.
       We count progress only up to the last point before
  the matching indicates that the ego is off route.

        \item \textbf{\scshape Pre-Lane Change.}
          For each pre-lane change evaluation, we multiply
          $P_{\mathrm{PLC}}$ by $0.7$ for a score of $0$
          and by $0.9$ for a score of $0.5$.
          Unencountered evaluation stop lines incur no pre-lane change penalty.

      \item \textbf{\scshape SD-route compliance.}
    We set $P_{\mathrm{SD}}=0$ if the rollout terminates because
  the ego exceeds the route-distance threshold defined above,
  or if post-rollout HMM matching identifies travel outside
  the designated route.
      If neither condition holds and matching succeeds,
      $P_{\mathrm{SD}}=1$.
      Near the start and end of a rollout, limited trajectory
      coverage around a junction can make connected roads difficult
      to distinguish. We allow an $8.5\,\mathrm{m}$ tolerance for
      such boundary matches, excluding travel opposite to the
      route direction.

    \item \textbf{\scshape Collision penalty.}
      We penalize at-fault collisions with pedestrians, vehicles,
      and bicycles using nuPlan-style criteria~\citep{nuplan1},
      excluding stationary-ego and active rear collisions.
      Front collisions and collisions with stationary agents count,
      with lateral collisions counted only when the ego spans
      multiple lanes or extends outside drivable areas.
      For each counted collision, we multiply $P_{\mathrm{col}}$
      by $0.5$ for pedestrians and $0.6$ for vehicles or
      bicycles~\citep{bench2drive}.
      If no collision is counted, $P_{\mathrm{col}}=1$.
      The same object is counted again only after contact breaks
      and either displacement exceeds $5\,\mathrm{m}$ or elapsed
      time exceeds $5\,\mathrm{s}$.

      \item \textbf{\scshape Off-road penalty.}
         We compute $P_{\mathrm{off}}=1-D_{\mathrm{off}}/D_{\mathrm{total}}$,
  where $D_{\mathrm{total}}$ is the total traveled distance
  and $D_{\mathrm{off}}$ is the distance driven outside
  drivable areas or against the direction of traffic.
  Driving-direction violations are not counted within intersections.

      \item \textbf{\scshape Traffic-light penalty.}
      Each red-light violation multiplies $P_{\mathrm{TL}}$ by $0.7$.
  A violation occurs when the ego occupies an intersection
  roadblock whose traffic signal is red.
  Each violation is counted once, even if it persists across
  multiple evaluation steps.

  \end{itemize}

  \subsection{Additional Driving Metrics}
We adapt Efficiency and Comfortness metrics from Bench2Drive~\citep{bench2drive} to our simulation setting, as detailed below.

  % \paragraph{\scshape Success Rate.}
  % A scenario succeeds
  % if the ego reaches its destination within the time limit
  % without any driving infractions.
  % Success Rate is the mean success indicator,
  % reported as a percentage.

  \paragraph{\scshape Efficiency.}
  Efficiency measures whether the ego maintains driving speed
  relative to other traffic.
  We compare its speed with vehicles that exceed $1\,\mathrm{m/s}$
  at least once during the rollout.
  At each evaluation step containing these vehicles, we compute
  their average speed. Efficiency is $100$ times the sum of ego
  speeds divided by the sum of these average vehicle speeds
  over the same steps.
  A score of $100$ indicates equal average speeds.

  \begin{minipage}[t]{0.5\linewidth}
\paragraph{\scshape Comfortness.}
  \noindent
      \vspace{0pt}
      Comfortness evaluates whether the ego drives smoothly
      without abrupt acceleration, braking, or steering.
      It is one if all limits in the table are satisfied
      throughout the trajectory, and zero otherwise.
      Except for longitudinal acceleration, the limits
      apply to absolute values.
  \end{minipage}
  \hfill
  \begin{minipage}[t]{0.48\linewidth}
      \vspace{-10pt}
      \centering
      \setlength{\tabcolsep}{4pt}
      \renewcommand{\arraystretch}{0.95}
      \resizebox{\linewidth}{!}{%
      \begin{tabular}{@{}lc@{}}
          \toprule
          \textbf{Comfort Criterion} & \textbf{Limit} \\
            \specialrule{0.3pt}{0.0ex}{0.15ex}
          Longitudinal acceleration
          & $[-4.05,\,2.40]\,\mathrm{m/s^2}$ \\
          Lateral acceleration
          & $4.89\,\mathrm{m/s^2}$ \\
          Yaw rate
          & $0.95\,\mathrm{rad/s}$ \\
          Yaw acceleration
          & $1.93\,\mathrm{rad/s^2}$ \\
          Longitudinal jerk
          & $4.13\,\mathrm{m/s^3}$ \\
          Derivative of acceleration magnitude
          & $8.37\,\mathrm{m/s^3}$ \\
          \bottomrule
      \end{tabular}%
      }
  \end{minipage}

\section{SD-Route Conditioning of Baseline Planners}
\label{app:sdroute}

This section details the model-specific adaptations of the baselines in
Section~\ref{main:sd_baseline}. We keep each original architecture and
training recipe, and add only the route encoder and the modules that read
it. We first transform the route points into the ego frame, then
  encode their positions relative to the first route point.
  Headings in the ego frame are encoded by their cosine and sine. The added
cross-attention has its own normalization and residual connection. Unavailable route points are masked, and a learned token
  represents the absence of route guidance when all route
  points are masked.

\begin{itemize}
  \item \textbf{LTF.} We use the camera-only latent variant. The
  trajectory query attends to the route in all three decoder layers,
  while the 30 agent queries do not. The ego status keeps only velocity
  and acceleration. Trained for 100 epochs.

  \item \textbf{DiffusionDrive.} The 20 anchor queries attend to the route
  in both decoder layers after the ego-query cross-attention. The route
  therefore conditions each of the two denoising steps. We use the
  camera-only latent variant. Trained for 100 epochs.

  \item \textbf{DrivoR.} The 64 proposal tokens attend to the route in all
  four refinement blocks. The ego history encoding is reduced from 11 to
  7 channels (pose, velocity, and acceleration). Trained for 25 epochs.

  \item \textbf{SafeDrive.} Every layer of both planning decoders attends
  to the same route tokens from a shared encoder. The ego status is
  reduced from 8 to 4 channels. We use the camera-only variant with
  pedestrian detection.

  \item \textbf{ReCogDrive.} A two-stage point encoder with max pooling
  compresses the 120 route points into one token. A Fourier embedding of
  the route start is added to this token, and the result is added to the
  ego-status embedding of the diffusion planner. The navigation line of
  the VLM prompt is generated from the same route, naming the next
  maneuver with its distance and the one after it (e.g., ``turn left in
  30\,m, then follow the road''). When the remaining route is too short
  to describe, the driving command is used instead.
\end{itemize}

\section{Additional Analysis}
\label{app:additional_results}

% \begin{table}[t]
%   \small
%   \centering
%   \setlength{\tabcolsep}{6pt}
%   \renewcommand{\arraystretch}{0.95}
%   \caption{Quantitative comparison with reconstruction baselines.
%   }
%   \label{tab:fixer_comparison}

%   \begin{tabular}{@{}lccc@{}}
%       \toprule
%       \textbf{Method}
%       & \textbf{PSNR} $\uparrow$
%       & \textbf{SSIM} $\uparrow$
%       & \textbf{LPIPS} $\downarrow$ \\
%       \midrule[0.3pt]

%       % OmniRe\citep{omnire}
%       % & 18.78 & 0.638 & 0.480 \\
%       % MTGS\citep{mtgs}
%       % & 18.94 & 0.677 & 0.439 \\
%       % HUGSIM\citep{hugsim}
%       % & 19.55 & 0.682 & 0.453 \\
%       % \midrule[0.3pt]

%       % \textbf{Ours}
%       % & \textbf{21.25} & \textbf{0.690} & \textbf{0.403} \\
%       OmniRe \citep{omnire}
%       & 19.4 & 0.654 & 0.456 \\
%       MTGS \citep{mtgs}
%       & 19.6 & 0.692 & 0.419 \\
%       HUGSIM \citep{hugsim}
%       & 20.1 & 0.701 & 0.436 \\
%       \midrule[0.3pt]

%       \textbf{Ours}
%       & \textbf{21.7} & \textbf{0.707} & \textbf{0.388} \\

%       \bottomrule
%   \end{tabular}
% \end{table}

\begin{table*}[!h]
  \centering
  \begin{minipage}[t]{0.58\textwidth}
      \centering
      \caption{The five nuPlan logs used for the comparison.
      Each scene is a 20\,s window sampled at 2\,Hz; frame
      indices are 2\,Hz keyframe indices within each log.}
      \vspace{1pt}
      \label{tab:scenes}
      \renewcommand{\arraystretch}{1.1}
      \setlength{\tabcolsep}{5pt}

      \resizebox{\linewidth}{!}{%
      \begin{tabular}{l l c c}
          \toprule[1pt]
          & \textbf{nuPlan log}
          & \textbf{City}
          & \textbf{Frames} \\
          \midrule[0.5pt]

          S1 & 2021.09.16.19.27.01\_veh-45\_01749\_03230
          & Pittsburgh & 1539--1578 \\

          S2 & 2021.09.29.19.02.14\_veh-28\_00273\_00514
          & Boston & 119--158 \\

          S3 & 2021.09.09.19.10.24\_veh-39\_01125\_01324
          & Boston & 120--159 \\

          S4 & 2021.06.28.21.16.05\_veh-14\_00957\_01198
          & Las Vegas & 270--309 \\

          S5 & 2021.06.28.20.24.43\_veh-38\_03385\_04952
          & Las Vegas & 1561--1600 \\

          \bottomrule[1pt]
      \end{tabular}%
      }
  \end{minipage}\hfill
  \begin{minipage}[t]{0.40\textwidth}
      \centering
      \caption{Quantitative comparison with reconstruction
      baselines. All baselines are retrained under their official
      recipes on identical data, and all metrics are computed
      with a single implementation.}
      \vspace{1pt}
      \label{tab:fixer_comparison}
      \renewcommand{\arraystretch}{1.1}
      \setlength{\tabcolsep}{5pt}

      \resizebox{\linewidth}{!}{%
      \begin{tabular}{l c c c}
          \toprule[1pt]
          \textbf{Method}
          & \textbf{PSNR} $\uparrow$
          & \textbf{SSIM} $\uparrow$
          & \textbf{LPIPS} $\downarrow$ \\
          \midrule[0.5pt]

          OmniRe~\citep{omnire}
          & 19.4 & 0.654 & 0.456 \\

          MTGS~\citep{mtgs}
          & 19.6 & 0.692 & 0.419 \\

          HUGSIM~\citep{hugsim}
          & 20.1 & 0.701 & 0.436 \\

          \midrule[0.5pt]

          \textbf{Ours}
          & \textbf{21.7} & \textbf{0.707} & \textbf{0.388} \\

          \bottomrule[1pt]
      \end{tabular}%
      }
  \end{minipage}
\end{table*}

\subsection{Comparison with SOTA 3DGS methods}
% We compare reconstruction quality with OmniRe~\citep{omnire},
% MTGS~\citep{mtgs}, and HUGSIM~\citep{hugsim} using PSNR,
% SSIM, and LPIPS.
% As shown in Table~\ref{tab:fixer_comparison}, our method
% achieves the highest PSNR and SSIM and the lowest LPIPS
% among the compared methods, indicating improvements in
% both image fidelity and perceptual similarity.
We compare reconstruction quality with OmniRe~\citep{omnire}, MTGS~\citep{mtgs}, and HUGSIM~\citep{hugsim} using PSNR, SSIM, and LPIPS.
We evaluate on five nuPlan~\citep{nuplan1} scenes from different logs across three cities, listed in
Table~\ref{tab:scenes}.
Each scene comprises a 20\,s window sampled at 2\,Hz, using all eight surround-view cameras at
$1920\times1080$ resolution.
Images are undistorted using the original intrinsic matrices as the output camera intrinsics.
We hold out every fourth keyframe across all eight cameras to evaluate novel-view synthesis at held-out camera poses.
We train each baseline from scratch using its official training recipe and iteration count.
All methods use the same camera poses, calibration, undistorted images, train/test split, and image resolution.
As shown in Table~\ref{tab:fixer_comparison}, our method achieves the highest PSNR and SSIM and the lowest LPIPS among the compared methods, indicating improvements in both image fidelity and perceptual similarity.

\subsection{Detailed RouteDS Analysis}
  Table~\ref{tab:sdroute_detailed} reports route completion
  and individual penalty factors alongside \textit{RouteDS}.
  SD-route guidance improves route compliance and \textit{RouteDS}
  across planners, but improvements in individual penalty
  factors are not consistent.
  These results distinguish gains in route following
  from changes in other driving criteria.

  \begin{table*}[!htbp]
      \centering
      \caption{Detailed \textit{RouteDS} results under non-reactive and
      reactive traffic. 
      }
      \label{tab:sdroute_detailed}
      \renewcommand{\arraystretch}{0.95}
      \setlength{\tabcolsep}{3pt}

      \resizebox{1.0\textwidth}{!}{%
      \begin{tabular}{l c | c c c c c c c | c c c c c c c}
          \toprule[1.3pt]

          \multirow[c]{2}{*}[-0.5ex]{\textbf{Method}}
          & \multirow[c]{2}{*}[-0.5ex]{%
              \shortstack[c]{\textbf{SD Route}\\\textbf{guidance}}}
          & \multicolumn{7}{c|}{\textbf{Non-reactive}}
          & \multicolumn{7}{c}{\textbf{Reactive}} \\
          \cmidrule(lr){3-9}
          \cmidrule(lr){10-16}

          &
          & \textit{\textbf{RouteDS}}
          & \textbf{RC$_{\mathrm{SD}}$}
          & \textbf{P$_{\mathrm{SD}}$}
          & \textbf{P$_{\mathrm{col}}$}
          & \textbf{P$_{\mathrm{off}}$}
          & \textbf{P$_{\mathrm{TL}}$}
          & \textbf{P$_{\mathrm{PLC}}$}
          & \textit{\textbf{RouteDS}}
          & \textbf{RC$_{\mathrm{SD}}$}
          & \textbf{P$_{\mathrm{SD}}$}
          & \textbf{P$_{\mathrm{col}}$}
          & \textbf{P$_{\mathrm{off}}$}
          & \textbf{P$_{\mathrm{TL}}$}
          & \textbf{P$_{\mathrm{PLC}}$} \\
          \midrule[0.85pt]

          \multicolumn{16}{c}{\textit{Vision-based planners}} \\
          \midrule[0.3pt]

          \multirow{2}{*}{LTF}
          & $\times$
          & 16.24 & 0.72 & 0.51 & 0.54 & 0.67 & 0.98 & 0.88
          & 16.27 & 0.75 & 0.54 & 0.59 & 0.67 & 0.98 & 0.86 \\
          & $\checkmark$
          & 20.13 & \textbf{0.99} & \underline{0.95} & 0.43 & 0.63 & 0.99 & 0.81
          & 23.01 & \textbf{0.99} & \underline{0.93} & 0.48 & 0.63 & 0.98 & 0.80 \\
          \midrule[0.3pt]

          \multirow{2}{*}{DiffusionDrive}
          & $\times$
          & 25.64 & 0.75 & 0.59 & 0.61 & 0.71 & \textbf{0.99} & \textbf{0.92}
          & 24.17 & 0.73 & 0.59 & 0.61 & 0.71 & 0.98 & 0.91 \\
          & $\checkmark$
          & 35.93 & \underline{0.96} & \textbf{1.00} & 0.51 & 0.76 & \textbf{0.99} & \underline{0.91}
          & 40.97 & \underline{0.98} & \textbf{1.00} & 0.57 & 0.76 & \underline{0.99} & 0.91 \\
          \midrule[0.3pt]

          \multirow{2}{*}{DrivoR}
          & $\times$
          & 20.61 & 0.68 & 0.50 & 0.69 & 0.86 & \textbf{0.99} & 0.90
          & 25.46 & 0.71 & 0.53 & 0.73 & 0.86 & 0.98 & 0.90 \\
          & $\checkmark$
          & \underline{36.45} & 0.84 & 0.90 & 0.65 & 0.86 & \underline{0.99} & 0.88
          & \underline{41.35} & 0.83 & 0.89 & 0.74 & 0.88 & 0.98 & 0.88 \\
          \midrule[0.3pt]

          \multirow{2}{*}{SafeDrive}
          & $\times$
          & 27.52 & 0.61 & 0.41 & \textbf{0.86} & \textbf{0.89} & 0.99 & \textbf{0.92}
          & 30.42 & 0.64 & 0.45 & \textbf{0.88} & \underline{0.89} & 0.98 & \textbf{0.92} \\
          & $\checkmark$
          & \textbf{44.35} & 0.85 & 0.82 & \underline{0.75} & \underline{0.89} & \underline{0.99} & 0.89
          & \textbf{48.94} & 0.84 & 0.79 & \underline{0.85} & \textbf{0.89} & 0.98 & 0.88 \\
          \midrule[0.85pt]

          \multicolumn{16}{c}{\textit{VLA-based planners}} \\
          \midrule[0.3pt]

          \multirow{2}{*}{ReCogDrive}
          & $\times$
          & 15.42 & 0.57 & 0.71 & 0.59 & 0.79 & \textbf{0.99} & 0.91
          & 20.33 & 0.53 & 0.62 & 0.70 & 0.80 & \underline{0.99} & \underline{0.91} \\
          & $\checkmark$
          & 21.45 & 0.54 & 0.83 & 0.69 & 0.83 & \textbf{0.99} & 0.90
          & 26.25 & 0.56 & 0.84 & 0.77 & 0.83 & \textbf{0.99} & 0.91 \\

          \bottomrule[1.3pt]
      \end{tabular}%
      }
  \end{table*}

\subsection{Additional Qualitative Analysis}
% In addition to the qualitative results below, we provide a \textcolor{red}{\textbf{supplementary video}} showing long-horizon closed-loop rollouts, comparisons between directional commands and explicit SD route guidance, and planner behavior across diverse driving scenarios.

\begin{figure}[h]
  \centering
  \includegraphics[width=0.9\textwidth]{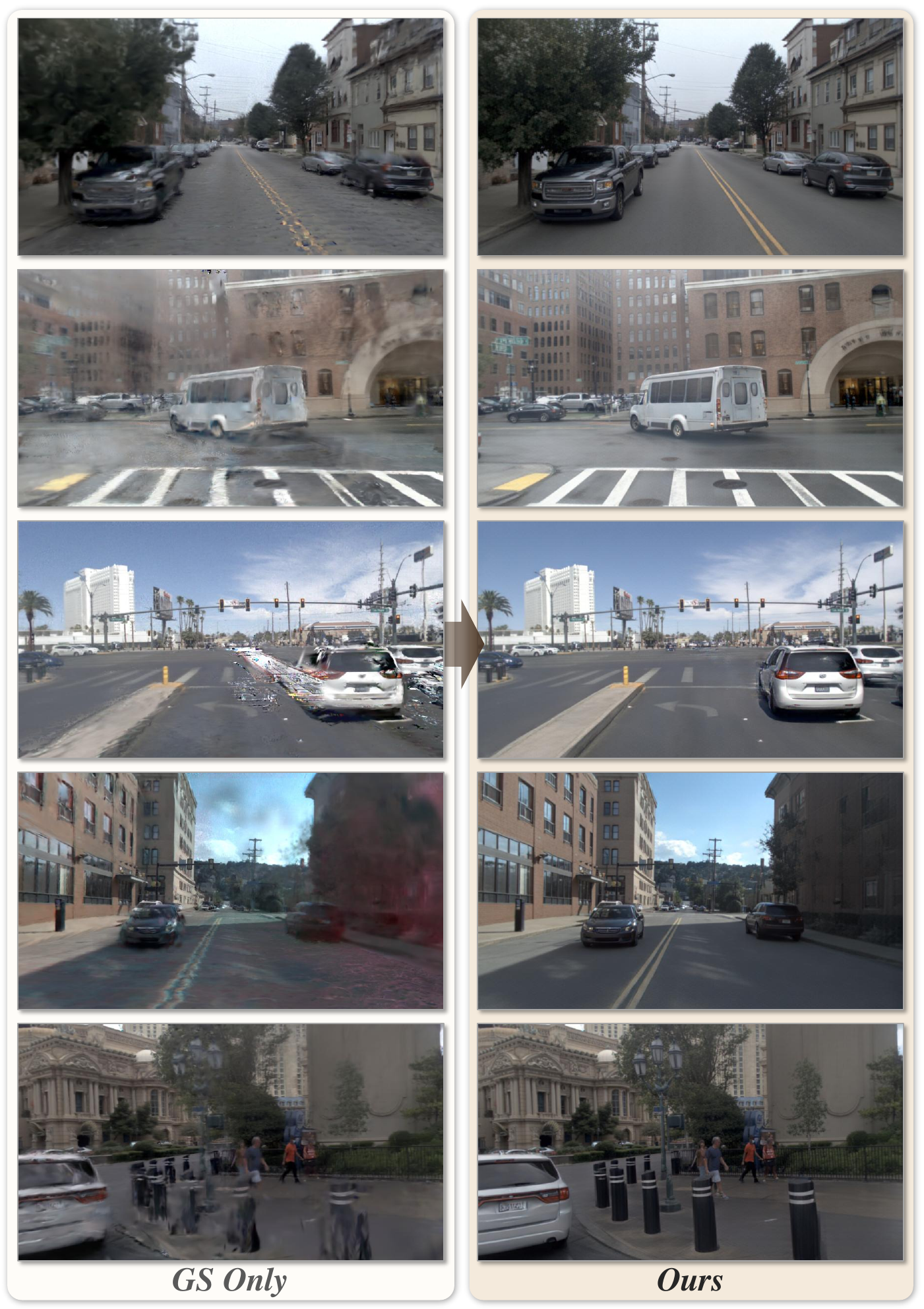}
  \caption{\textbf{Qualitative Comparison of Fixer Refinement.} We show challenging novel views with limited coverage from training observations, where 3DGS-only renderings exhibit pronounced artifacts. Fine-tuned Fixer reduces these artifacts and improves visual detail.}
  \label{fig:fixer_appendix}
\end{figure}

\paragraph{\scshape Novel-view refinement.}
Figure~\ref{fig:fixer_appendix} shows novel views shifted by $3$--$5\,\mathrm{m}$ from the recorded camera positions, where limited observational coverage leads to pronounced reconstruction artifacts.
Fine-tuned Fixer reduces blur, distortions, ghosting, and color and brightness inconsistencies across diverse scene elements, including lane markings, vehicles, pedestrians, and background structures.

\begin{figure}[t]
  \centering
  \includegraphics[width=\textwidth]{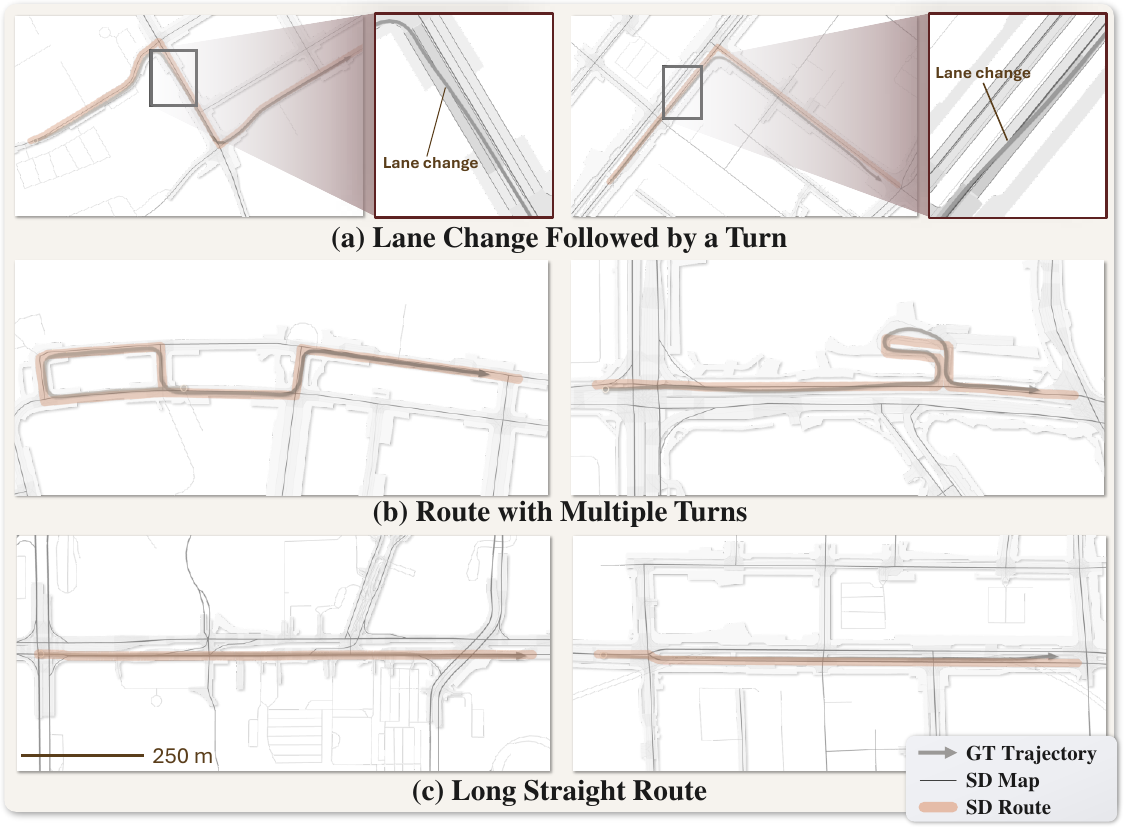}
  \caption{\textbf{Example trajectories and SD-Routes from Odyssey.} The benchmark includes lane changes in preparation for upcoming turns, routes with multiple turns, and long straight routes. These scenarios span varying route complexity and include anticipatory maneuvers that reflect the driver's navigation intent.}
  \label{fig:scenario_appendix}
\end{figure}

\paragraph{\scshape Scenario diversity.}
Figure~\ref{fig:scenario_appendix} overlays recorded ego
trajectories and SD routes for scenarios involving lane changes
before turns, multiple turns, and long straight routes.
These examples span different levels of route complexity.
In particular, lane changes before turns illustrate how
navigation intent shapes driving decisions before the
maneuver itself, motivating evaluation over extended sequences.

\begin{figure}[t]
  \centering
  \includegraphics[width=\textwidth]{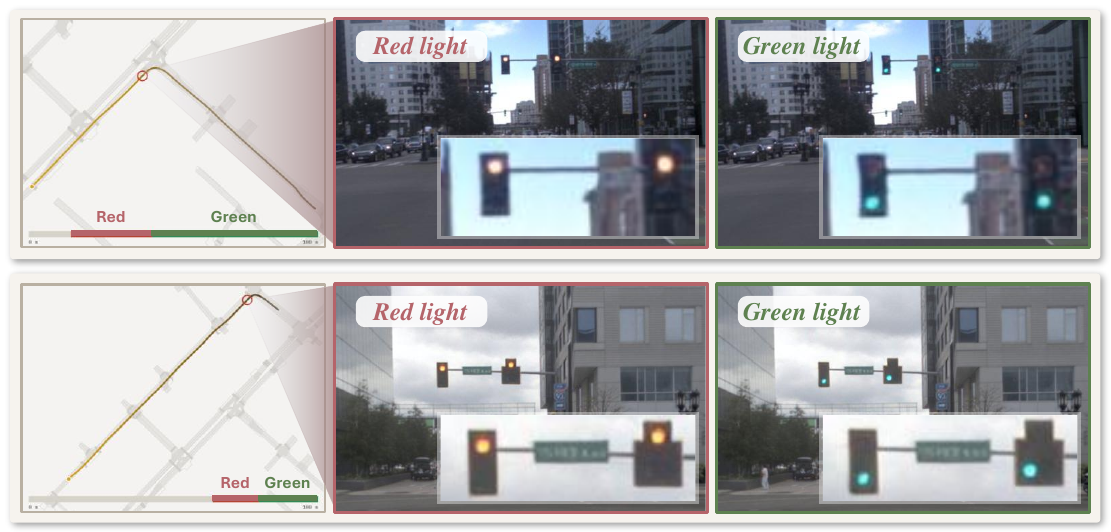}
  \caption{\textbf{Additional Qualitative Results of Traffic Light Modeling.} Our model reproduces red-to-green transitions in sync with the traffic-light schedule in each scene.}
  \label{fig:tl_appendix}
\end{figure}

\paragraph{\scshape Traffic-light modeling.}
Figure~\ref{fig:tl_appendix} shows additional
examples of reconstructed traffic-light states alongside
their locations and signal schedules.
The enlarged views highlight red and green states reproduced
by the model, illustrating its ability to represent signal
changes within the same reconstructed scene.

% \subsection{Qualitative Analysis of 3DGS}

% \subsection{Ablation Study for 3D GS}
% \section{Evaluation Scenarios}
% \subsection{Data Curation}
% \subsection{Scenario Distributions}

\end{document}